\documentclass[mathpazo]{cmr}

\usepackage{bm}

\usepackage[a4paper,
            left=1in,
            right=1in,
            top=1in,
            bottom=1in,
            ]{geometry}

\begin{document}
%%%%% title : short title may not be used but TITLE is required.
% \title{TITLE}
% \title[short title]{TITLE}
\title{Convergence analysis of kernel learning FBSDE filter}

%%%%% author(s):%%%%%%%%%%%%%
% Single author:
% \author[name in running head]{AUTHOR\corrauth}
% [name in running head] is NOT OPTIONAL, it is a MUST.
% Use \corrauth to indicate the corresponding author.
% Use \email to provide email address of author.
% Use \address to provide complete address with post code of the institution where the work was done.
% \author[Yunzheng Lyu]{Author\corrauth}
% \address{School of Mathematics,
%          Jilin University,
%          Changchun, Jilin, 130012, P.R. China. } %complete address with Post Code
% \email{{\tt author@email} (O.~Author)}

%% Multiple authors:
%% Note the use of \affil and \affilnum to link names and addresses.
%% The author for correspondence is marked by \corrauth.
%% use \emails to provide email addresses of authors.
%% e.g. below example has 3 authors, first author is also the corresponding author,
%%      author 1 and 3 having the same address.
\author[Yunzheng Lyu, Feng Bao]{ Yunzheng Lyu\affil{1}\comma\corrauth,
                                       Feng Bao\affil{1} }
\address{\affilnum{1}\ Department of Mathematics,
                      Florida State University,
                      FL, 32306, USA \\}
\emails{{\tt yl19@fsu.edu} (Yunzheng Lyu),
       {\tt fbao@fsu.edu}  (Feng Bao)}

% \footnote and \thanks are not used in the heading section.
% Another acknowlegments/support of grants, state in Acknowledgments section
% \section*{Acknowledgments}

%%%%% Begin Abstract %%%%%%%%%%%
\begin{abstract}
Kernel learning forward backward SDE filter is an iterative and adaptive meshfree approach to solve the nonlinear filtering problem. It builds from forward backward SDE for Fokker-Planker equation, which defines evolving density for the state variable, and employs KDE to approximate density. This algorithm has shown more superior performance than mainstream particle filter method, in both convergence speed and efficiency of solving high dimension problems. 

However, this method has only been shown to converge empirically. In this paper, we present a rigorous analysis to demonstrate its local and global convergence, and provide theoretical support for its empirical results. 
\end{abstract}
%%%%%  end %%%%%%%%%%%

%%%%% AMS/Keywords %%%%%%%%%%%
    %AMS subject classification can be found in http://mathscinet.ams.org/msc/msc2010.html
    %List 3 to 8 keywords

%%%% maketitle %%%%%
\maketitle

%%%%%% Start %%%%%%%%%%%%%%%%%
\section{Introduction}

Optimal filtering problem is to estimate unknown underlying state variables $S_{t}$ from associated noisy observation data $O_{t}$. In the general application setting, prior distribution is known for the underlying state system, combined with likelihood function relating state variables to observation data, one can perform Bayesian inference to estimate evolving posterior density for the unknown state. It has applications in many fields, ranging from signal processing, quantum physics, mathematical finance to machine learning and AI etc.

When both the state and observation systems are linear, Kalman filter provides optimal analytical solution for posterior state density (Kalman and Bucy [1]). To cope with nonlinear filtering problems, different variations of Kalman filter have been proposed. Extended Kalman filter linearly approximates nonlinear transformations by applying tangent linear operator or Jacobian matrix (Julier and Uhlmann [2]). Ensemble Kalman filter generates ensemble of samples to approximate Covariance for nonlinear system (Evensen [3]). These variations within the Kalman filter framework perform badly when non-linearity is high.

Following the work of bootstrap filter (Gordon, Salmon and Smith [4]), Particle filter (Del Moral [5]) or sequential Monte Carlo (Liu and Chen [6]) simulate posterior state density by propagating samples through a sequence of importance sampling, resampling and MCMC steps. Various improvements have been made later to refine these intermediate sub-steps, e.g. Auxiliary Particle filter (Pitt and Shephard [7]) and Population Monte Carlo (Iba [8]). This framework can handle nonlinear system effectively, but it suffers degeneracy issue in long term or high frequency simulations.

Another line of work calculates posterior state density analytically through SPDE (Kallianpur and Striebel [9]; Zakai [10]). Their major drawback is slow convergence and profound complexity, especially for high-dimensional problems. Due to the equivalence found between FBDSDE and certain parabolic SPDE (Pardoux, Peng
[11]), FBDSDE framework is established to estimate posterior state density ([12]-[18]). Archibald and Bao ([19]) further simplifies this framework to FBSDE filter algorithm, in which only prior state density is derived through the FBSDE system, sample posterior density is then approximated through Bayesian inference and posterior density function is learnt from sample results by kernel learning methods. This framework is mesh free and can deal with high-dimensional problems efficiently.

 Along the development of algorithms, convergence analysis of filtering algorithms is more subtle. Crisan and Doucet ([20]) prove the convergence of a general particle filter under suitable regularity conditions. In this paper, we follow similar convergence analysis framework and show the convergence results for FBSDE filter.

 The rest of paper is organized as follows. In section 2, we summarize the theoretical background for FBSDE filter. In section 3, we demonstrate implementations of FBSDE filter algorithm. In section 4, we conduct convergence analysis for the algorithm. And concluding remarks are given in section 5.

\section{FBSDE Filter}
Underlying problems in FBSDE filters take stochastic system form:
\begin{equation}\tag{2.1}\label{eq2.1}
\begin{aligned}
dS_{t}&=g(S_{t})dt+\sigma_{t}dW_{t}\\
dO_{t}&=h(S_{t})dt+(r_{t})dV_{t}
\end{aligned}
\end{equation}

where $\int_{0}^{t}\sigma_{s}dW_{s} =w_{t} \sim N(0,Q_{t})$, $S_t \in R^{d_{x}}$ and $O_t \in R^{d_{y}}$.

FBSDE filter predicts prior state variable density $p(S_{t}|O_{t^-})$ ($O_{t^-}$ is observation data right before time t) through a system of FBSDE equations and then updates posterior density $p(S_{t}|O_{t})$ for state variable by Bayesian inference.

From underlying problem (\ref{eq2.1}), we can construct a FBSDE equation:
\begin{equation}\label{eq2.2}\tag{2.2}
\begin{aligned}
\overline{X}_{t} &=x +\int_{0}^{t}g(\overline{ X}_{s})ds+\int_{0}^{t}\sigma_{s} dW_{s}\\
\overline{Y}_{t} &= \psi(\overline{X}_{T})-\int_{t}^{T}\overline{Z}_{s} dW_{s}
\end{aligned}
\end{equation}

Taking expectation for backward SDE in (\ref{eq2.2}), we can get:
$$\overline{Y}_{t}=\overline{u}(t,x)=E[\psi(\overline{X}_{T})|\overline{X}_{t}=x]$$

By Feymann-Kac formula, $\overline{u}(t,x)$ is solution to following PDE:
\begin{equation*}\label{formula3.1}
\begin{aligned}
-\frac{\partial \overline{u}}{\partial t}=\sum_{j=1}^{d_{x}}g_{j}(x)\frac{\partial \overline{u}}{\partial x_{j}}+\frac{1}{2}\sum_{i=1}^{d_{x}}\sum_{j=1}^{d_{x}}(\sigma_{t}\sigma_{t}^{T})_{ij}\frac{\partial^{2} \overline{u}}{\partial x_{i}\partial x_{j}}, \quad \overline{u}(T,x)=\psi(x)
\end{aligned}
\end{equation*}

By setting terminal condition $\psi(x)$ as indicator function $1_{A}(S_{T})$, where $A\in \mathcal{B}(R^{d_{x}})$, this restricted form of Feymann-Kac PDE is equivalent to Kolmogorov backward equation.

And the corresponding Kolmogorov forward/Fokker-Planck equation is:
$$\frac{\partial u}{\partial t}=-\sum_{j=1}^{d_{x}}\frac{\partial g_{j}(x)u}{\partial x_{j}}+\frac{1}{2}\sum_{i=1}^{d_{x}}\sum_{j=1}^{d_{x}}(\sigma_{t}\sigma_{t}^{T})_{ij}\frac{\partial^{2} u}{\partial x_{i}\partial x_{j}} \quad u(0,x)=p_{0}(S_{0}=x)$$

By generalization of Feynman-Kac formula (Pardoux, Peng 
 [11]), we can derive the corresponding FBSDE for Fokker-Planck equation:
\begin{equation}\label{eq2.3}\tag{2.3}
\begin{aligned}
X_{t} &=x -\int_{t}^{T}g(X_{s})ds+\int_{t}^{T}\sigma_{s} d\overleftarrow{W}_{s}\\
Y_{t} &= p_{0}(X_{0})-\int_{0}^{t} \sum_{j=1}^{d_{x}}\frac{\partial g_{j}}{\partial x_{j}}(X_{s})Y_{s}ds-\int_{0}^{t} Z_{s} d\overleftarrow{W}_{s}
\end{aligned}
\end{equation}

$Y_{t}$ in (\ref{eq2.3}) is $u(t,x)$ in Fokker-Planck equation, and defines evolving state variable density $p(S_{t})$ forward in time. 

State variable density $Y_{t}$ in (\ref{eq2.3}) evolves independently with observation data, and can be used to propagate prior density for state variable.

From Fokker-Planck equation, we can also conclude that (\ref{eq2.3}) only holds for time-dependent $\sigma_{t}$, if we use $\sigma_{t}(X_{t})$, forward SDE in (\ref{eq2.3}) shall be:
\begin{equation*}\label{eq2.4}\tag{2.4}
\begin{aligned}
Y_{t} &= p_{0}(X_{0})-\int_{0}^{t} \sum_{j=1}^{d_{x}}\frac{\partial g_{j}}{\partial x_{j}}(X_{s})Y_{s}ds-\int_{0}^{t} Z_{s}
d\overleftarrow{W}_{s}\\
&\quad +\int_{0}^{t}\frac{1}{2}\sum_{i=1}^{d_{x}}\sum_{j=1}^{d_{x}}[\frac{\partial^2 (\sigma_{t}\sigma_{t}^{T})_{ij}}{\partial x_{i}\partial x_{j}}Y_{s}+2\frac{\partial (\sigma_{t}\sigma_{t}^{T})_{ij}}{\partial x_{i}}\frac{\partial Y_{s}}{\partial x_{j}}]ds
\end{aligned}
\end{equation*}

Corresponding numerical scheme for (\ref{eq2.3}) is:
\begin{equation}\label{eq2.5}\tag{2.5}
\begin{aligned}
X_{k-1} &= X_{k}-g(X_{k})\Delta t_{k-1} + \sigma_{t_{k}} \Delta W_{t_{k-1}}\\
Y_{k}^{O_{t_{k-1}}}(X_{k}) &= E_{t_{k}}^{X_{k}}[Y_{k-1}(X_{k-1})]-\sum_{j=1}^{d_{x}}\frac{\partial g_{j}}{\partial x_{j}}(X_{k})Y_{k}^{O_{t_{k-1}}}(X_{k}) \Delta t_{k-1}\\
\end{aligned}
\end{equation}

where $E_{t_{k}}^{X_{k}}[Y_{k-1}(X_{k-1})]=E[Y_{k-1}(X_{k-1})|X_{k},O_{t_{k-1}}]$

Posterior state variable density estimate $Y_{k}(X_{k})$ for $p(S_{t_{k}}=X_{k}|O_{t_{k}})$ can be updated by Bayesian inference:
$$Y_{k}(X_{k})=\frac{p(O_{t_{k}}|S_{k}=X_{k})Y_{k}^{O_{t_{k-1}}}(X_{k})}{p(O_{t_{k}}|O_{t_{k-1}})}$$

\section{FBSDE Filter Algorithm}
In filtering algorithm descriptions, we use common notation $x_{t_{k}}$ for true variable values and $x_{k}$ for estimated variable results, for corresponding variable $x$ at time $t_{k}$.

FBSDE filter algorithm follows the standard prediction and update framework, where prediction of prior density for state variable is propagated with FBSDE equations and update of posterior density is achieved by Bayesian inference. It also resamples more heavily in high density value region to prevent sample degeneration in long time simulation.

(1) Initialization:
\begin{quote}
    Set k=1, sample N times from initial state variable distribution:
    $$X_{0}^{i}\sim p_{0}(S_{0})dS_{0} \quad i=1,\dots,N$$
    The corresponding estimated posterior pdf for X at time 0 is exact:
    $$Y_{0}^{N,M}(x)=p_{0}(x)$$
\end{quote}

(2) Prediction:
\begin{quote}
    For each sample $i=1,\dots,N$:\\
    (2.a) Propagate state variable samples forward in time by (\ref{eq2.1}):
    $$\widetilde{X}_{k}^{i}=X_{k-1}^{i}+g(X_{k-1}^{i})(t_{k}-t_{k-1})+\sigma_{t_{k-1}}\Delta W_{t_{k-1}}^{i}$$
    
    (2.b) Set initial value for fixed point iteration: $\widetilde{Y}_{k}^{O_{t_{k-1}},i,0}=Y_{k-1}^{N,M}(\widetilde{X}_{k}^{i})$.\\
    For $m=1,\cdots,M$:\\
    Propagate backward to obtain $\widetilde{X}_{k-1}^{i,m}$ by (\ref{eq2.5}):
    $$\widetilde{X}_{k-1}^{i,m} = \widetilde{X}_{k}^{i}-g(\widetilde{X}_{k}^{i})\Delta t_{k-1} + \sigma_{t_{k}} \Delta W_{t_{k-1}}^{i,m}$$
    Then estimate $\widetilde{Y}_{k}^{O_{t_{k-1}},i,m}$ by fixed point iteration:
    \begin{equation*}
    \begin{aligned}
    \widetilde{Y}_{k}^{O_{t_{k-1}},i,m} &= E_{t_{k}}^{\widetilde{X}_{k}^{i},m}[Y_{k-1}^{N,M}(\widetilde{X}_{k-1})]-\sum_{j=1}^{d_{x}}\frac{\partial g_{j}}{\partial x_{j}}(\widetilde{X}_{k}^{i})\widetilde{Y}_{k}^{O_{t_{k-1}},i,m-1}\Delta t_{k-1}\\
    \end{aligned}
    \end{equation*}
    where
    \begin{equation*}
    \begin{aligned}
    E_{t_{k}}^{\widetilde{X}_{k}^{i},m}[Y_{k-1}^{N,M}(\widetilde{X}_{k-1})]&=E[Y_{k-1}^{N,M}(\widetilde{X}_{k-1})|\widetilde{X}_{k}^{i},O_{t_{k-1}}]\\
    &=\frac{1}{m}\sum_{j=1}^{m}Y_{k-1}^{N,M}(\widetilde{X}_{k-1}^{i,j})
    \end{aligned}
    \end{equation*}
    
    (2.c) Obtain empirical prior density estimation:
    $$\widetilde{Y}_{k}^{O_{k-1},N,M}(x)=\widetilde{Y}_{k}^{O_{t_{k-1}},i,M} \quad  \mbox{if } \quad  x=\widetilde{X}_{k}^{i}$$
\end{quote}

(3) Update:
\begin{quote}
    Use Bayesian inference to update posterior state variable density estimate:
    $$\widetilde{Y}_{k}^{i,M}=\frac{p(O_{t_{k}}|S_{k}=\widetilde{X}_{k}^{i})\widetilde{Y}_{k}^{O_{t_{k-1}},i,M}}{\sum_{j=1}^{N}p(O_{t_{k}}|S_{k}=\widetilde{X}_{k}^{j})\frac{1}{N}} \quad i=1,\dots,N$$
    where $O_{t_{k}}|S_{k} \sim N(O_{t_{k-1}}+\int_{t_{k-1}}^{t_{k}}{h(S_t)dt}|S_{k},t_{k}-t_{k-1})$\\
    And empirical posterior density estimate is:
    \begin{equation*}
    \begin{aligned}
        \widetilde{Y}_{k}^{N,M}(x)=\widetilde{Y}_{k}^{i,M} \quad  \mbox{if } \quad  x=\widetilde{X}_{k}^{i}
        %\begin{cases}
        %    \widetilde{Y}_{k}^{i} & \mbox{if } %x=\widetilde{X}_{k}^{i}\\
        %    0 & otherwise
        %\end{cases}
    \end{aligned}
    \end{equation*}
    So the posterior density for state variable depends on both sample size N and fixed point iteration step M.
\end{quote}

(4) Kernel function $Y_{k}^{N,M}(x)$ learning by KDE
\begin{quote}
    (4.a) From training data $\{(\widetilde{X}_{k}^{i},\widetilde{Y}_{k}^{i,M})\}_{i=1}^{N}$, choose L ($L<<N$) kernel centers $\{\widehat{X}_{k}^{l}\}_{l=1}^{L}$, construct kernel function $\widehat{Y}_{t_{k}}(x)$:
    $$\widehat{Y}_{t_{k}}^{L}(x)=\sum_{l=1}^{L}\alpha^{l}_{t_{k}}\phi(x|\widehat{X}_{k}^{l},\lambda_{t_{k}}^{l})$$
    where $\phi(x|\widehat{X}_{k}^{l},\lambda_{t_{k}}^{l})=\exp(-\frac{|x-\widehat{X}_{k}^{l}|^{2}}{(\lambda_{t_{k}}^{l})^{2}})$
    
    (4.b) Learn parameters $\{\alpha^{l}_{t_{k}},\lambda_{t_{k}}^{l}\}_{l=1}^{L}$ by stochastic gradient descent:\\
    Define loss function:
    $$L_{t_{k}}(\alpha,\lambda)=\frac{1}{N}\sum_{i=1}^{N}L_{t_{k}}(\alpha,\lambda,i)$$
    $$L_{t_{k}}(\alpha,\lambda,i)=(\widehat{Y}^{L}_{t_{k}}(\widetilde{X}_{k}^{i})-\widetilde{Y}_{k}^{i,M})^2$$
    For step $s=1,\cdots,S$:\\
    Choose a random sample index i(s) from $1,\cdots,N$;\\
    Learn $\{\alpha^{l}_{t_{k}},\lambda_{t_{k}}^{l}\}_{l=1}^{L}$ by gradient descent with prechosen initialization $\alpha^{l}(k,0)$ and $\lambda^{l}(k,0)$:
    \begin{equation*}
    \begin{aligned}
        \alpha^{l}(k,s)&=\alpha^{l}(k,s-1)-\rho_{\alpha_{t_{k}}^{l}}^{s}\frac{\partial L_{t_{k}}(\alpha,\lambda,i(s))}{\partial \alpha_{t_{k}}^{l}}\\
        \lambda^{l}(k,s)&= \lambda^{l}(k,s-1)-\rho_{ \lambda_{t_{k}}^{l}}^{s}\frac{\partial L_{t_{k}}(\alpha,\lambda,i(s))}{\partial  \lambda_{t_{k}}^{l}}\\
    \end{aligned}    
    \end{equation*}
    where
    \begin{equation*}
    \begin{aligned}
        \frac{\partial L_{t_{k}}(\alpha,\lambda,i(s))}{\partial \alpha_{t_{k}}^{l}}
        &=2(\widehat{Y}_{t_{k}}^{L}(\widetilde{X}_{k}^{i(s)})-\widetilde{Y}_{k}^{i(s),M})\phi(\widetilde{X}_{k}^{i(s)}|\widehat{X}_{k}^{l},\lambda^{l}(k,s-1))\\
        \frac{\partial L_{t_{k}}(\alpha,\lambda,i(s))}{\partial \lambda_{t_{k}}^{l}}
        &=2(\widehat{Y}_{t_{k}}^{L}(\widetilde{X}_{k}^{i(s)})-\widetilde{Y}_{k}^{i(s),M})\alpha^{l}(k,s-1)\\
        &\quad *\phi(\widetilde{X}_{k}^{i(s)}|\widehat{X}_{k}^{l},\lambda^{l}(k,s-1))\frac{2|\widetilde{X}_{k}^{i(s)}-\widehat{X}_{k}^{l}|^{2}}{(\lambda^{l}(k,s-1))^{3}}
    \end{aligned}    
    \end{equation*}
    $\rho_{\alpha_{t_{k}}^{l}}^{s}$ and $\rho_{ \lambda_{t_{k}}^{l}}^{s}$ are learning rates, and can be varied based on algorithm step $k$, gradient descent iteration step $s$ and different parameter index $l$.\\
    
    (4.c) Obtain trained kernel function $Y_{k}^{N,M}(x)$:
    $$Y_{k}^{N,M}(x)=\sum_{l=1}^{L}\alpha^{l}(k,S)\phi(x|\widehat{X}_{k}^{l},\lambda^{l}(k,S))$$
    Note that $Y_{k}^{N,M}(\widetilde{X}_{k}^{i})$ may be different from empirical posterior estimate $\widetilde{Y}_{k}^{i}$. And we will use $Y_{k}^{N,M}(\widetilde{X}_{k}^{i})$ as posterior pdf estimate in resampling and next iteration for $\widetilde{X}_{k}^{i}$.
\end{quote}

(5) Resampling:
\begin{quote}
    For $i=1,\dots, N$:\\
    set $X_{k}^{i}=\widetilde{X}_{k}^{i}$;\\
    generate a new random sample from  $Y_{k}^{N,M}(x)$:
    $$X_{k}^{i,new}\sim Y_{k}^{N,M}(x)dx$$
    reset $X_{k}^{i}=X_{k}^{i,new}$ with probability:
    $$min{\{1,\frac{Y_{k}^{N,M}(X_{k}^{i,new})}{Y_{k}^{N,M}(X_{k}^{i})}\}}$$
    Now we have N new samples:
    $$\{X_{k}^{i}\}_{i=1}^{N}$$
    And the corresponding posterior density estimate is:
    $$Y_{k}^{N,M}(x)=\sum_{l=1}^{L}\alpha^{l}(k,S)\phi(x|\widehat{X}_{k}^{l},\lambda^{l}(k,S))$$.
\end{quote}

(6) k=k+1, repeat steps (2)-(6) until $k>\mathbb{K}$ (i.e. $k=\mathbb{K}+1$).

FBSDE filter avoids sample degeneration since it can compute weights directly without referring to resampling, it also retains high computation efficiency due to KDE kernel learning of state variable density.

Disadvantages of FBSDE filter include constraints on the form of underlying filtering problem. $\sigma_{t}$ in (\ref{eq2.1}) has to be deterministic, otherwise FBSDE equation needs to include derivatives of $\sigma_{t}$ with respect to state variables, as shown in (\ref{eq2.4}). $r_{t}$ in (\ref{eq2.1}) can be incorporated through Bayesian inference, and therefore can even be stochastic.

\section{Convergence Analysis for KDE kernel learning FBSDE filter}

\subsection{KDE kernel learning convergence analysis}\label{sec41}

Kernel density estimation (KDE) is also called Parzen–Rosenblatt window method.

Given i.i.d samples $(x_{1},x_{2},\cdots,x_{n})$ from unknown density f(x), for any x $\in$ $R^{d_{x}}$, KDE estimates f(x) by\\
$$\hat{f}_{n}(x)=\frac{1}{n}\sum_{i=1}^{n}K_{h}(x-x_{i})=\frac{1}{nh^{d_{x}}}\sum_{i=1}^{n}K(\frac{x-x_{i}}{h})$$

where K is kernel, a non-negative integrable function, and $K_{h}(x)=\frac{1}{h^{d_{x}}}K(\frac{x}{h})$. $h>0$ is called bandwidth and can be used to adjust domain from [-1,1] to [-h, h].

For 1-dimension case $x\in R^{1}$, KDE convergence rate with respect to sample size $n$ is $n^{-\frac{4}{5}}$ (Wahba [21]). Following the convergence proof of one-dimension case, we can extend results to multi-dimension case $x\in R^{d_{x}}$:

Define $\alpha=(\alpha_{1},\cdots,\alpha_{d_{x}})$ ($\alpha_{i}\geq 0, \quad 1\leq i\leq d_{x}$): 

$|\alpha|=\sum_{i=1}^{d_{x}}|\alpha_{i}|$;

$x^{\alpha}=x_{1}^{\alpha_{1}} \cdots x_{d_{x}}^{\alpha_{d_{x}}}$

$f^{(\alpha)}(x)=\frac{\partial^{|\alpha|} f}{\partial x_{1}^{\alpha_{1}}\cdots\partial x_{d_{x}}^{\alpha_{d_{x}}}}(x)$

$W_{p}^{(m)}$: Sobolev space of functions whose first m-1 derivatives are absolutely continuous and whose mth derivative is in $L^{p}$.

$W_{p}^{(m)}(M)=\{f:f\in W_{p}^{(m)},||f^{(m)}||_{p}\leq M\}$

\begin{assumption}\label{assumption411}

Kernel K(x) is a real-valued integrable function on $R^d_{x}$ satisfying:\\

(i) $\displaystyle \sup_{x\in R^{d_{x}}}|K(x)|<\infty$

(ii) $\lim_{|x|\rightarrow\infty} |x|K(x)=0$ where $|x|=\sum_{i=1}^{d_{x}}|x_{i}|$

(iii) $\int_{R^{d_{x}}}|K(x)|dx<\infty$ and $\int_{R^{d_{x}}}K(x)dx=1$

(iv) $\int_{R^{d_{x}}}x^{\alpha}K(x)dx=0$ where $|\alpha|=1,2,...,m-1$

(v) $\int_{R^{d_{x}}}|x|^{m}|K(x)|dx<\infty$ 

(vi) $\lim_{n\rightarrow \infty}h=0$ and $\lim_{n\rightarrow \infty}nh^{d_{x}}=\infty$
\end{assumption}

\begin{theorem}\label{theorem412}

Let p be integer, $p\geq1$. Let $f\in W_{p}^{(m)}(M)$ and $\sup_{x\in R^{d_{x}}}f(x)\leq \Lambda$.

Kernel K satisfies properties (i)-(vi) and $$\hat{f}_{n}(x)=\frac{1}{n}\sum_{i=1}^{n}K_{h}(x-x_{i})=\frac{1}{nh^{d_{x}}}\sum_{i=1}^{n}K(\frac{x-x_{i}}{h})$$

where $h=[\frac{Bd_{x}}{2mnM^2A}]^{\frac{1}{2m+d_{x}}}$

Then for $\forall x\in R^{d_{x}}$:
$$E[(f(x)-\hat{f}_{n}(x))^2]\leq Dn^{-\phi(2,p)}(1+o(1))$$

with $\phi(2,p)=\frac{2m}{2m+d_{x}}$, which is irrelevant with p

$D=\theta(M^2A)^{\frac{d_{x}}{2m+d_{x}}}(B)^{\frac{2m}{2m+d_{x}}}$

$\theta=\frac{2m+d_{x}}{(4)^{\frac{2m}{2m+d_{x}}}(d_{x})^{\frac{d_{x}}{2m+d_{x}}}}$

$A=\frac{1}{[(m-1)!]
^{2}[(m-1)q+1]^{\frac{2}{q}}}[\int_{R^{d_{x}}} |K(y)||y|^{m}dy]^{2}$

$B=\Lambda \int_{-\infty}^{\infty}K^2(x)dx$
\end{theorem}

\vspace{12pt}
Proof:

Define $$f_n(x)=E[\hat{f}_{n}(x)]=\int_{R^{d_{x}}}\frac{1}{h^{d_{x}}}K(\frac{x-y}{h})f(y)dy$$

The squared error at given sample $x$ can be decomposed into a variance term and a bias term:
$$E[(\hat{f}_{n}(x)-f(x))^2]=E[(\hat{f}_{n}(x)-f_{n}(x))^2]+E[({f}_{n}(x)-f(x))^2]$$

From [2], we can generalize theorem 2A to multi-dimensional state variable space with constant bandwidth $h$:
$\lim_{n\rightarrow \infty} nh^{d_{x}}Var(\hat{f}_{n}(x))=f(x)\int_{R^{d_{x}}}K^2(y)dy$. And then variance term is:
\begin{equation*}
\begin{aligned}
E[(\hat{f}_{n}(x)-f_{n}(x))^2]
&= \frac{1}{nh^{d_{x}}}(f(x)\int_{R^{d_{x}}}K^{2}(y)dy-h^{d_{x}}(f(x)\int_{R^{d_{x}}}K(y)dy)^2+o(1))\\
&\leq B\frac{1}{nh^{d_{x}}}(1+o(1))
\end{aligned}
\end{equation*}

where $B=\Lambda \int_{R^{d_{x}}}K^{2}(y)dy$.

The bias term is:
$$E[({f}_{n}(x)-f(x))^2]=({f}_{n}(x)-f(x))^2=(\int_{R^{d_{x}}}K(-y)(f(x+yh)-f(x))dy)^2$$

Applying Taylor's theorem in multi-variables:
\begin{equation*}
\begin{aligned}
f(x+yh)-f(x)&=\int_{0}^{1}\frac{df(x+yht_{1})}{dt}dt_{1}\\
&=\sum_{|\alpha|=1}\frac{\partial^{(\alpha)}f(x)}{\alpha!}(yh)^{\alpha}+\int_{0}^{1}\int_{0}^{t_{1}}\frac{d^{2}f(x+yht_{2})}{dt^{2}}dt_{2}dt_{1}\\
&=\sum_{|\alpha|=1}^{m-1}\frac{\partial^{(\alpha)}f(x)}{\alpha!}(yh)^{\alpha}+\int_{0}^{1}\int_{0}^{t_{1}}\cdots\int_{0}^{t_{n-1}}\frac{d^{m}f(x+yht_{m})}{dt^{m}}dt_{m}\cdots dt_{2}dt_{1}\\
&=\sum_{|\alpha|=1}^{m-1}\frac{\partial^{(\alpha)}f(x)}{\alpha!}(yh)^{\alpha}+\int_{0}^{1}\frac{d^{m}f(x+yht_{m})}{dt^{m}}\frac{(1-t_{m})^{m-1}}{(m-1)!}dt_{m}\\
&=\sum_{|\alpha|=1}^{m-1}\frac{\partial^{(\alpha)}f(x)}{\alpha!}(yh)^{\alpha}+m\int_{0}^{1}\sum_{|\alpha|=m}\frac{\partial^{\alpha}f(x+yht)}{\alpha!}(yh)^{\alpha}(1-t)^{m-1}dt\\
\end{aligned}
\end{equation*}

With assumptions (iii)-(v) in \ref{assumption411} and an application of Holder inequality, the bias term can be simplified and bounded:
\begin{equation*}
\begin{aligned}
|f_{n}(x)-f(x)|&=|\int_{R^{d_{x}}}K(-y)(f(x+yh)-f(x))dy|\\
&=|\int_{R^{d_{x}}}K(-y)mh^{m}\sum_{|\alpha|=m}\frac{y^{\alpha}}{\alpha!}\int_{0}^{1}\partial^{\alpha}f(x+yht)(1-t)^{m-1}dtdy|\\
&\leq \frac{h^{m}}{(m-1)!}\int_{R^{d_{x}}} |K(y)||y|^{m}dy\frac{M}{((m-1)q+1)^{\frac{1}{q}}}
\end{aligned}
\end{equation*}
where $\frac{1}{p}+\frac{1}{q}=1$

Combining bias term and variance term, we have:
$$E[(\hat{f}_{n}(x)-f(x))^2]<=Ah^{2m}M^2+B\frac{1}{nh^{d_{x}}}(1+o(1))$$
where $A=\frac{1}{[(m-1)!]
^{2}[(m-1)q+1]^{\frac{2}{q}}}[\int_{R^{d_{x}}} |K(y)||y|^{m}dy]^{2}$ and $B=\Lambda \int_{R^{d_{x}}}K^{2}(y)dy$

Ignoring o(1) factor, RHS can be minimized at $h=[\frac{Bd_{x}}{2mnAM^{2}}]^{\frac{1}{2m+dx}}$, and minimal RHS is $\frac{2m+d_{x}}{(d_{x})^{\frac{d_{x}}{2m+d_{x}}}(2m)^{\frac{2m}{2m+d_{x}}}}(AM^{2})^{\frac{d_{x}}{2m+d_{x}}}(B)^{\frac{2m}{2m+d_{x}}}n^{-\frac{2m}{2m+d_{x}}}(1+o(1))$

\vspace{12pt}

For Gaussian kernels, $m=2$. So from theorem \ref{theorem412}, we can conclude that for multi-dimensional state variable $x\in R^{d_{x}}$, Gaussian KDE converges with a polynomial order of $\frac{4}{4+d_{x}}$ to true underlying density. Hence for FBSDE filter, we have:

\begin{equation*}\tag{4.1.1}\label{eq411}
\begin{aligned}
\lim_{L \rightarrow \infty}\widehat{Y}_{t_{k}}^{L}(x)=\lim_{N \rightarrow \infty}\overline{Y}_{k}^{N,M}(x) \quad \forall x \in R^{d_{x}}, M \in R
\end{aligned}
\end{equation*}

Because of this pointwise convergence, loss function in kernel learning converges to global minimum 0. But stochastic gradient descent may not reach global minimum due to local minimum or straddle points.

We claim that loss function in Gaussian kernel learning will almost surely converge to the global minimum 0, by showing loss function is asymptotically convex.

\begin{theorem}\label{theorem413}
Hessian matrix for $L_{t_{k}}(\alpha,\lambda,i)$ is asymptotically positive semi-definite in convex set $R^{d_{x}}$, therefore $L_{t_{k}}(\alpha,\lambda,i)$ is asymptotically convex.
\end{theorem}

\vspace{12pt}
Proof:

The first order of derivative of $L_{t_{k}}(\alpha,\lambda,i)$ is:
\begin{equation}
\begin{aligned}
    \frac{\partial L_{t_{k}}(\alpha,\lambda,i)}{\partial \alpha_{t_{k}}^{l}}
    &=2(\widehat{Y}_{t_{k}}^{L}(\widetilde{X}_{k}^{i})-\widetilde{Y}_{k}^{i})\phi^{l}_{t_{k}}\\
    \frac{\partial L_{t_{k}}(\alpha,\lambda,i)}{\partial \lambda_{t_{k}}^{l}}
    &=2(\widehat{Y}_{t_{k}}^{L}(\widetilde{X}_{k}^{i})-\widetilde{Y}_{k}^{i})\alpha^{l}_{t_{k}}
    \phi^{l}_{t_{k}}A^{l}_{t_{k}}
\end{aligned}
\end{equation}

where $\phi^{l}_{t_{k}}=\phi(\widetilde{X}_{k}^{i}|\widehat{X}_{k}^{l},\lambda^{l}_{t_{k}})$
and $A^{l}_{t_{k}}=\frac{2|\widetilde{X}_{k}^{i}-\widehat{X}_{k}^{l}|^{2}}{(\lambda^{l}_{t_{k}})^{3}}$

The second order of derivative of $L_{t_{k}}(\alpha,\lambda,i)$ is:
\begin{equation*}
\begin{aligned}
    \frac{\partial^{2} L_{t_{k}}(\alpha,\lambda,i)}{\partial \alpha_{t_{k}}^{l}\partial \alpha_{t_{k}}^{m}}
    &=2\phi^{l}_{t_{k}}\phi^{m}_{t_{k}}\\
    \frac{\partial^{2} L_{t_{k}}(\alpha,\lambda,i)}{\partial \alpha_{t_{k}}^{l}\partial \alpha_{t_{k}}^{l}}
    &=2(\phi^{l}_{t_{k}})^{2}\\
    \frac{\partial L_{t_{k}}(\alpha,\lambda,i)}{\partial \lambda_{t_{k}}^{l}\partial \lambda_{t_{k}}^{m}}
    &=2\alpha^{l}_{t_{k}}
    \phi^{l}_{t_{k}}A^{l}_{t_{k}}\alpha^{m}_{t_{k}}
    \phi^{m}_{t_{k}}A^{m}_{t_{k}}\\
    \frac{\partial L_{t_{k}}(\alpha,\lambda,i)}{\partial \lambda_{t_{k}}^{l}\partial \lambda_{t_{k}}^{l}}
    &=2(\alpha^{l}_{t_{k}}
    \phi^{l}_{t_{k}}A^{l}_{t_{k}})^{2}+2(\widehat{Y}_{t_{k}}^{L}(\widetilde{X}_{k}^{i})-\widetilde{Y}_{k}^{i})\alpha^{l}_{t_{k}}
    \phi^{l}_{t_{k}}((A^{l}_{t_{k}})^{2}-\frac{6|\widetilde{X}_{k}^{i}-\widehat{X}_{k}^{l}|^{2}}{(\lambda^{l}_{t_{k}})^{4}})\\
    \frac{\partial^{2} L_{t_{k}}(\alpha,\lambda,i)}{\partial \alpha_{t_{k}}^{l}\partial \lambda_{t_{k}}^{m}}
    &=2\phi^{l}_{t_{k}}\alpha^{m}_{t_{k}}
    \phi^{m}_{t_{k}}A^{m}_{t_{k}}\\
    \frac{\partial^{2} L_{t_{k}}(\alpha,\lambda,i)}{\partial \alpha_{t_{k}}^{l}\partial \lambda_{t_{k}}^{l}}
    &=2(\phi^{l}_{t_{k}})^2\alpha^{l}_{t_{k}}
    A^{l}_{t_{k}}+2(\widehat{Y}_{t_{k}}^{L}(\widetilde{X}_{k}^{i})-\widetilde{Y}_{k}^{i})
    \phi^{l}_{t_{k}}A^{l}_{t_{k}}\\
\end{aligned}
\end{equation*}

From convergence of kernel density estimation, we know when kernel size L goes to infinity, $\widehat{Y}_{t_{k}}^{L}(\widetilde{X}_{k}^{i})$ will converge to $\widetilde{Y}_{k}^{i}$, therefore the items containing $\widehat{Y}_{t_{k}}^{L}(\widetilde{X}_{k}^{i})-\widetilde{Y}_{k}^{i}$ will disappear, and asymptotic Hessian matrix of $L_{t_{k}}(\alpha,\lambda,i)$ is:

\[
\begin{aligned}
H&=
 \begin{bmatrix}
 \frac{\partial^{2} L_{t_{k}}(\alpha,\lambda,i)}{\partial \alpha_{t_{k}}^{1}\partial \alpha_{t_{k}}^{1}} & \frac{\partial^{2} L_{t_{k}}(\alpha,\lambda,i)}{\partial \alpha_{t_{k}}^{1}\partial \alpha_{t_{k}}^{2}} & \cdots & 
 \frac{\partial^{2} L_{t_{k}}(\alpha,\lambda,i)}{\partial \alpha_{t_{k}}^{1}\partial \lambda_{t_{k}}^{L-1}} &
 \frac{\partial^{2} L_{t_{k}}(\alpha,\lambda,i)}{\partial \alpha_{t_{k}}^{1}\partial \lambda_{t_{k}}^{L}}\\
 
  \frac{\partial^{2} L_{t_{k}}(\alpha,\lambda,i)}{\partial \alpha_{t_{k}}^{2}\partial \alpha_{t_{k}}^{1}} & \frac{\partial^{2} L_{t_{k}}(\alpha,\lambda,i)}{\partial \alpha_{t_{k}}^{2}\partial \alpha_{t_{k}}^{2}} & \cdots & 
 \frac{\partial^{2} L_{t_{k}}(\alpha,\lambda,i)}{\partial \alpha_{t_{k}}^{2}\partial \lambda_{t_{k}}^{L-1}} &
 \frac{\partial^{2} L_{t_{k}}(\alpha,\lambda,i)}{\partial \alpha_{t_{k}}^{2}\partial \lambda_{t_{k}}^{L}}\\
 
   \vdots & \vdots &\ddots & \vdots & \vdots\\
     
   \frac{\partial^{2} L_{t_{k}}(\alpha,\lambda,i)}{\partial \alpha_{t_{k}}^{L}\partial \alpha_{t_{k}}^{1}} & \frac{\partial^{2} L_{t_{k}}(\alpha,\lambda,i)}{\partial \alpha_{t_{k}}^{L}\partial \alpha_{t_{k}}^{2}} & \cdots & 
  \frac{\partial^{2} L_{t_{k}}(\alpha,\lambda,i)}{\partial \alpha_{t_{k}}^{L}\partial \lambda_{t_{k}}^{L-1}} &
   \frac{\partial^{2}
   L_{t_{k}}(\alpha,\lambda,i)}{\partial \alpha_{t_{k}}^{L}\partial \lambda_{t_{k}}^{L}}\\
   
   \frac{\partial^{2} L_{t_{k}}(\alpha,\lambda,i)}{\partial \lambda_{t_{k}}^{1}\partial \alpha_{t_{k}}^{1}} & \frac{\partial^{2} L_{t_{k}}(\alpha,\lambda,i)}{\partial \lambda_{t_{k}}^{1}\partial \alpha_{t_{k}}^{2}} & \cdots & 
  \frac{\partial^{2} L_{t_{k}}(\alpha,\lambda,i)}{\partial \lambda_{t_{k}}^{1}\partial \lambda_{t_{k}}^{L-1}} &
   \frac{\partial^{2}
   L_{t_{k}}(\alpha,\lambda,i)}{\partial \lambda_{t_{k}}^{1}\partial \lambda_{t_{k}}^{L}}\\

    \vdots & \vdots &\ddots & \vdots & \vdots\\
     
   \frac{\partial^{2} L_{t_{k}}(\alpha,\lambda,i)}{\partial \alpha_{t_{k}}^{L-1}\partial \alpha_{t_{k}}^{1}} & \frac{\partial^{2} L_{t_{k}}(\alpha,\lambda,i)}{\partial \alpha_{t_{k}}^{L-1}\partial \alpha_{t_{k}}^{2}} & \cdots & 
  \frac{\partial^{2} L_{t_{k}}(\alpha,\lambda,i)}{\partial \alpha_{t_{k}}^{L-1}\partial \lambda_{t_{k}}^{L-1}} &
   \frac{\partial^{2}
   L_{t_{k}}(\alpha,\lambda,i)}{\partial \alpha_{t_{k}}^{L-1}\partial \lambda_{t_{k}}^{L}}\\
   
   \frac{\partial^{2} L_{t_{k}}(\alpha,\lambda,i)}{\partial \lambda_{t_{k}}^{L}\partial \alpha_{t_{k}}^{1}} & \frac{\partial^{2} L_{t_{k}}(\alpha,\lambda,i)}{\partial \lambda_{t_{k}}^{L}\partial \alpha_{t_{k}}^{2}} & \cdots & 
  \frac{\partial^{2} L_{t_{k}}(\alpha,\lambda,i)}{\partial \lambda_{t_{k}}^{L}\partial \lambda_{t_{k}}^{L-1}} &
   \frac{\partial^{2}
   L_{t_{k}}(\alpha,\lambda,i)}{\partial \lambda_{t_{k}}^{L}\partial \lambda_{t_{k}}^{L}}\\
\end{bmatrix}
\end{aligned}
\]

\begin{footnotesize}
\[
\begin{aligned}
&=2
\begin{bmatrix}
(\phi^{1}_{t_{k}})^2 & \phi^{1}_{t_{k}}\phi^{2}_{t_{k}} &\cdots &  \phi^{1}_{t_{k}}\alpha^{L-1}_{t_{k}}
    \phi^{L-1}_{t_{k}}A^{L-1}_{t_{k}} & \phi^{1}_{t_{k}}\alpha^{L}_{t_{k}}
    \phi^{L}_{t_{k}}A^{L}_{t_{k}}\\
    
    \phi^{2}_{t_{k}}\phi^{1}_{t_{k}} & (\phi^{2}_{t_{k}})^{2} &\cdots &   \phi^{2}_{t_{k}}\alpha^{L-1}_{t_{k}}
    \phi^{L-1}_{t_{k}}A^{L-1}_{t_{k}} & \phi^{2}_{t_{k}}\alpha^{L}_{t_{k}}
    \phi^{L}_{t_{k}}A^{L}_{t_{k}}\\
    
    \vdots & \vdots &\ddots & \vdots & \vdots\\
    
    \phi^{L}_{t_{k}}\phi^{1}_{t_{k}} & \phi^{L}_{t_{k}}\phi^{2}_{t_{k}} &\cdots &  \phi^{L}_{t_{k}}\alpha^{L-1}_{t_{k}}
    \phi^{L-1}_{t_{k}}A^{L-1}_{t_{k}} & (\phi^{L}_{t_{k}})^{2}\alpha^{L}_{t_{k}}
    A^{L}_{t_{k}}\\
    
    \alpha^{1}_{t_{k}}
    A^{1}_{t_{k}}(\phi^{1}_{t_{k}})^{2} & \alpha^{1}_{t_{k}}\phi^{1}_{t_{k}}
    A^{1}_{t_{k}}\phi^{2}_{t_{k}} &\cdots &  \alpha^{1}_{t_{k}}\phi^{1}_{t_{k}}
    A^{1}_{t_{k}}\alpha^{L-1}_{t_{k}}\phi^{L-1}_{t_{k}}
    A^{L-1}_{t_{k}}&
    \alpha^{1}_{t_{k}}\phi^{1}_{t_{k}}
    A^{1}_{t_{k}}\alpha^{L}_{t_{k}}\phi^{L}_{t_{k}}
    A^{L}_{t_{k}}\\
    
    \vdots & \vdots &\ddots & \vdots & \vdots\\
    
    \alpha^{L-1}_{t_{k}}
    A^{L-1}_{t_{k}}\phi^{L-1}_{t_{k}}\phi^{1}_{t_{k}}& \alpha^{L-1}_{t_{k}}\phi^{L-1}_{t_{k}}
    A^{L-1}_{t_{k}}\phi^{2}_{t_{k}} &\cdots &   (\alpha^{L-1}_{t_{k}}\phi^{L-1}_{t_{k}}
    A^{L-1}_{t_{k}})^{2}&
    \alpha^{L-1}_{t_{k}}\phi^{L-1}_{t_{k}}
    A^{L-1}_{t_{k}}\alpha^{L}_{t_{k}}\phi^{L}_{t_{k}}
    A^{L}_{t_{k}}\\
    
    \alpha^{L}_{t_{k}}
    A^{L}_{t_{k}}\phi^{L}_{t_{k}}\phi^{1}_{t_{k}}& \alpha^{L}_{t_{k}}\phi^{L}_{t_{k}}
    A^{L}_{t_{k}}\phi^{2}_{t_{k}} &\cdots &  \alpha^{L}_{t_{k}}\phi^{L}_{t_{k}}
    A^{L}_{t_{k}}\alpha^{L-1}_{t_{k}}\phi^{L-1}_{t_{k}}
    A^{L-1}_{t_{k}}&
    (\alpha^{L}_{t_{k}}\phi^{L}_{t_{k}}
    A^{L}_{t_{k}})^{2}\\
\end{bmatrix}
\end{aligned}
\]
\end{footnotesize}

H is 2L*2L matrix. One significant property of this matrix is that 2*2 matrix formed by intersection of two adjacent rows and any two columns (adjacent or not) has determinant 0. 

There are three cases. The first one contains two adjacent rows from first L rows, which is between gradient vectors for $\frac{\partial L_{t_{k}}(\alpha,\lambda,i)}{\partial \alpha_{t_{k}}}$ and represented by first two rows in H. The second case contains row L and L+1, which is between gradient vector for $\frac{\partial L_{t_{k}}(\alpha,\lambda,i)}{\partial \alpha^{L}_{t_{k}}}$ and $\frac{\partial L_{t_{k}}(\alpha,\lambda,i)}{\partial \lambda^{1}_{t_{k}}}$. It's represented by middle two rows in H. The third case contains two adjacent rows from last L rows, which is between gradient vectors for $\frac{\partial L_{t_{k}}(\alpha,\lambda,i)}{\partial \lambda_{t_{k}}}$ and represented by last two rows in H.

It's easy to verify that any 2*2 sub-matrix formed in the 3 cases above all have determinant 0. In fact, any 2*2 matrix formed by intersection of two random rows and two random columns have determinant 0.

Based on this property, we can deduce that the all upper left square matrix have determinant 0 (except first upper left 1*1 matrix which has only one positive element $\frac{\partial^{2} L_{t_{k}}(\alpha,\lambda,i)}{\partial (\alpha_{t_{k}}^{1}))^{2}}$), since determinant of N*N matrix can be expressed as a linear combination of determinants of (N-1)*(N-1) sub-matrix. Therefore, asymptotic Hessian matrix for $L_{t_{k}}(\alpha,\lambda,i)$ is PSD.

\vspace{18pt}

% Corollary: For loss function $L_{t_{k}}(\alpha,\lambda)=\frac{1}{B}\sum_{i=1}^{B} {L_{t_{k}}(\alpha,\lambda,i)}$, the asymptotic Hessian matrix, where is average of asymptotic Hessian matrix for each 
% sample loss ${L_{t_{k}}(\alpha,\lambda,i)}$, is also PSD. Therefore, loss function $L_{t_{k}}(\alpha,\lambda)$ is also asymptotically convex.

\begin{lemma}\label{lemma414}
Under assumption \ref{assumption411}, for $\forall x \in R^{d_{x}}, M \in R$,
$$\lim_{L \rightarrow \infty}Y_{k}^{N,M}(x)=\lim_{N \rightarrow \infty}\overline{Y}_{k}^{N,M}(x)$$
with sufficiently large number of gradient descent steps.

Convergence rate for RMSE is $O(L^{-\frac{2}{4+d_x}})$.
\end{lemma}

\vspace{18pt}
Proof:

From theorem \ref{theorem412} and theorem \ref{theorem413}, we know single sample loss function $L_{t_{k}}(\alpha,\lambda,i)$ is asymptotically convex and 0. For loss function $L_{t_{k}}(\alpha,\lambda)=\frac{1}{N}\sum_{i=1}^{N} {L_{t_{k}}(\alpha,\lambda,i)}$, when $L$ goes to infinity, both sample loss $L_{t_{k}}(\alpha,\lambda,i)$ and sample size $N$ are affected, so we can't simple apply the limit operator.

However, error bound for sample loss $L_{t_{k}}(\alpha,\lambda,i)$ is independent of x, and this uniform bound will also bound the average loss $L_{t_{k}}(\alpha,\lambda)$:
$$E[L_{t_{k}}(\alpha,\lambda)]=\frac{1}{N}\sum_{i=1}^{N}E[L_{t_{k}}(\alpha,\lambda,i)]\leq O(L^{-\frac{4}{4+d_x}})$$

Following same argument, Hessian matrix for average loss function $L_{t_{k}}(\alpha,\lambda)$ is also asymptotically PSD, so $L_{t_{k}}(\alpha,\lambda)$ is asymptotically convex and 0. 

Hence we can conclude:

\begin{equation*}\tag{4.1.2}\label{eq412}
\begin{aligned}
\lim_{L \rightarrow \infty} & L_{t_{k}}(\alpha,\lambda) =0\\
\lim_{L \rightarrow \infty} &Y_{k}^{N,M}(x)=\lim_{L \rightarrow \infty}\widehat{Y}_{t_{k}}^{L}(x) \quad \forall x \in R^{d_{x}}, M \in R
\end{aligned}
\end{equation*}

Combining equation (\ref{eq411}) and (\ref{eq412}), we have:
$$\lim_{L \rightarrow \infty}Y_{k}^{N,M}(x)=\lim_{N \rightarrow \infty}\overline{Y}_{k}^{N,M}(x) \quad \forall x \in R^{d_{x}}, M \in R$$

With sufficiently large number of gradient descent steps, the convergence rate for mean squared error is $O(L^{-\frac{4}{4+d_x}})$, since both KDE and Hessian matrix converge at this rate. But this convergence rate only serves as a reference. On one hand, this convergence rate is the optimal result for uniform bandwidth, since parameters in KDE are not unique when loss function reaches the global minimum 0. For example, when $h=[\frac{Bd_{x}}{2mnAM^{2}}]^{\frac{1}{2m+dx+1}}$, the loss function will also converge to 0 at a smaller convergence rate $O(L^{-\frac{4}{5+d_x}})$. On the other hand, the bandwidth may not be uniform, and the algorithm does propose different $\lambda$ for different samples, in which case the convergence rate can be faster than $O(L^{-\frac{4}{4+d_x}})$.

Because we use gradient descent to train the empirical kernel model $Y_{k}^{N,M}(x)$, we need to assume that number of gradient descent steps is sufficiently large, so that empirical model $Y_{k}^{N,M}$ will converge to the underlying KDE kernel model $\widehat{Y}_{t_{k}}^{L}(x)$. The convergence rate of gradient descent steps can be another aspect we further explore. But we skip this detail here because we can directly use kernel model parameters from theorem \ref{theorem412} and save the trouble of gradient descent steps convergence analysis.
\vspace{18pt}

% Remark: The loss function is defined based on batch size $B$, if instead, we define loss function with all sample points: $L_{t_{k}}(\alpha,\lambda)=\frac{1}{N}\sum_{i=1}^{N} {L_{t_{k}}(\alpha,\lambda,i)}$, when $L$ goes to infinity, both sample loss $L_{t_{k}}(\alpha,\lambda,i)$ and sample size $N$ go to infinity.

\subsection{Fixed point iteration convergence analysis}\label{sec42}

FBSDE filter uses step size M to combine fixed point itertaion and expectation estimation and estimate prior state variable density:
\begin{equation*}
\begin{aligned}
\widetilde{Y}_{k}^{O_{t_{k-1}},i,m} &= E_{t_{k}}^{\widetilde{X}_{k}^{i},m}[Y_{k-1}^{N}(\widetilde{X}_{k-1})]-\sum_{j=1}^{d_{x}}\frac{\partial g_{j}}{\partial x_{j}}(\widetilde{X}_{k}^{i})\widetilde{Y}_{k}^{O_{t_{k-1}},i,m-1}\Delta t_{k-1}\\
\end{aligned}
\end{equation*}

Convergence of fixed point iteration entails contraction mapping region and convergence rate will affect size of M.

But we can avoid convergence analysis for fixed point iteration by noticing that fixed point iteration comes from right point approximation for drift integral, if we use left point approximation instead, we replace $\widetilde{Y}_{k}^{O_{t_{k-1}}}$ by $\overline{Y}_{k}^{O_{t_{k-1}}}$:
\begin{equation*}
\begin{aligned}
\overline{Y}_{k}^{O_{t_{k-1}},i,M} &= E_{t_{k}}^{\widetilde{X}_{k}^{i},M}[Y_{k-1}^{N}(\widetilde{X}_{k-1})]\\
& \quad -\sum_{j=1}^{d_{x}}E_{t_{k}}^{\widetilde{X}_{k}^{i},M}[\frac{\partial g_{j}}{\partial x_{j}}(\widetilde{X}_{k-1})Y_{k-1}^{N}(\widetilde{X}_{k-1})]\Delta t_{k-1}\\
\end{aligned}
\end{equation*}

where $E_{t_{k}}^{\widetilde{X}_{k}^{i},M}[f(\widetilde{X}_{k-1})]=\frac{1}{M}\sum_{j=1}^{M}f(\widetilde{X}_{k-1}^{j,i})$

The empirical prior density estimate is:
\begin{equation*}
\begin{aligned}
    \overline{Y}_{k}^{O_{t_{k-1}},N,M}(x)=\overline{Y}_{k}^{O_{t_{k-1}},i,M} \quad \mbox{if } \quad x=\widetilde{X}_{k}^{i}
\end{aligned}
\end{equation*}

\begin{lemma} \label{lemma321}Under assumption that $|\sum_{j=1}^{d_{x}}\frac{\partial g_{j}}{\partial x_{j}}(x)|$ is bounded by $G$ and $Y_{k-1}^{N,M}(x)=p(S_{t_{k-1}}=x|O_{t_{k-1}})$, for $\forall x \in R^{d_x}$

$$\lim_{M \rightarrow \infty}\overline{Y}_{k}^{O_{t_{k-1}},N,M}(x)=p(S_{k}=x|O_{t_{k-1}})$$

where $S_{k}$ is a discrete time update of $S_{t_{k}}$:
$$S_{k}=S_{t_{k-1}}+g(S_{t_{k-1}})(t_{k}-t_{k-1})+\sigma_{t_{k-1}}\Delta W_{t_{k-1}}$$

and $p(S_{k}=x|O_{t_{k-1}})$ is prior pdf for $S_{k}$ based on $Y_{k-1}^{N,M}$:
\begin{equation*}
\begin{aligned}
p(S_{k}=x|O_{t_{k-1}})&= E_{t_{k}}^{x}[p(S_{t_{k-1}}=\widetilde{X}_{k-1}|O_{t_{k-1}})]\\
&-\sum_{j=1}^{d_{x}}E_{t_{k}}^{x}[\frac{\partial g_{j}}{\partial x_{j}}(\widetilde{X}_{k-1})p(S_{t_{k-1}}=\widetilde{X}_{k-1}|O_{t_{k-1}})]\Delta t_{k-1}\\
\end{aligned}
\end{equation*}

The uniform convergence rate of mean squared error is $\frac{1}{M}$.

\end{lemma}

\vspace{18pt}
Proof:

For $\forall x \in R^{d_{x}}:$

\begin{equation*}
\begin{aligned}
E&[(\overline{Y}_{k}^{O_{t_{k-1}},N,M}(x)-p(S_{k}=x|O_{t_{k-1}}))^2]\\
&=E[((E_{t_{k}}^{x,M}[Y_{k-1}^{N,M}(\widetilde{X}_{k-1})]-E_{t_{k}}^{x}[Y_{k-1}^{N,M}(\widetilde{X}_{k-1})])\\
&-(E_{t_{k}}^{x,M}[\sum_{j=1}^{d_{x}}\frac{\partial g_{j}}{\partial x_{j}}(\widetilde{X}_{k-1})Y_{k-1}^{N,M}(\widetilde{X}_{k-1})]\Delta t-E_{t_{k}}^{x}[\sum_{j=1}^{d_{x}}\frac{\partial g_{j}}{\partial x_{j}}(\widetilde{X}_{k-1})Y_{k-1}^{N,M}(\widetilde{X}_{k-1})]\Delta t))^2]\\
&\leq 2E[(E_{t_{k}}^{x,M}[Y_{k-1}^{N,M}(\widetilde{X}_{k-1})]-E_{t_{k}}^{x}[Y_{k-1}^{N,M}(\widetilde{X}_{k-1})])^2]\\
&+2E[(E_{t_{k}}^{x,M}[\sum_{j=1}^{d_{x}}\frac{\partial g_{j}}{\partial x_{j}}(\widetilde{X}_{k-1})Y_{k-1}^{N,M}(\widetilde{X}_{k-1})]\Delta t-E_{t_{k}}^{x}[\sum_{j=1}^{d_{x}}\frac{\partial g_{j}}{\partial x_{j}}(\widetilde{X}_{k-1})Y_{k-1}^{N,M}(\widetilde{X}_{k-1})]\Delta t)^2]\\
& =2\frac{Var_{t_{k}}^{x}[Y_{k-1}^{N,M}(\widetilde{X}_{k-1})]}{M}+2\Delta t^2\frac{Var_{t_{k}}^{x}[\sum_{j=1}^{d_{x}}\frac{\partial g_{j}}{\partial x_{j}}(\widetilde{X}_{k-1})Y_{k-1}^{N,M}(\widetilde{X}_{k-1})]}{M}\\
& \leq \frac{2(1+T^2G^2)Var_{t_{k}}^{x}[{Y_{k-1}^{N,M}(\widetilde{X}_{k-1})}]}{M}
\end{aligned}
\end{equation*}

% $Y_{k-1}^{N}$ is Gaussian density and bounded, so $Var_{t_{k}}^{x}[Y_{k-1}^{N}(\widetilde{X}_{k-1})]$ is bounded.

% $\sum_{j=1}^{d_{x}}\frac{\partial g_{j}}{\partial x_{j}}$ is assumed to be $L_4$ integrable, $Var_{t_{k}}^{x}[\sum_{j=1}^{d_{x}}\frac{\partial g_{j}}{\partial x_{j}}(\widetilde{X}_{k-1})Y_{k-1}^{N}(\widetilde{X}_{k-1})]$ can be shown to be finite:

% \begin{equation*}
% \begin{aligned}
% Var_{t_{k}}^{x}[\sum_{j=1}^{d_{x}}\frac{\partial g_{j}}{\partial x_{j}}(\widetilde{X}_{k-1})Y_{k-1}^{N}(\widetilde{X}_{k-1})]
% &\leq E_{t_{k}}^{x}[(\sum_{j=1}^{d_{x}}\frac{\partial g_{j}}{\partial x_{j}}(\widetilde{X}_{k-1})Y_{k-1}^{N}(\widetilde{X}_{k-1}))^2]\\
% &\leq \sqrt{E_{t_{k}}^{x}[(\sum_{j=1}^{d_{x}}\frac{\partial g_{j}}{\partial x_{j}}(\widetilde{X}_{k-1}))^4]}\sqrt{E_{t_{k}}^{x}[(Y_{k-1}^{N}(\widetilde{X}_{k-1}))^4]}\\
% &< \infty
% \end{aligned}
% \end{equation*}

$Y_{k-1}^{N,M}$ is Gaussian density and bounded, therefore error bound doesn't depend on x and convergence is uniform. Hence we can conclude:

$\lim_{M \rightarrow \infty}\overline{Y}_{k}^{O_{t_{k-1}},N,M}(x)=p(S_{k}=x|O_{t_{k-1}}) \quad a.e.$ with strong uniform convergence rate $\frac{1}{\sqrt{M}}$ for RMSE.

From kernel convergence analysis in theorem \ref{theorem412}, error bound $Var_{t_{k}}^{x}[{Y_{k-1}^{N,M}(\widetilde{X}_{k-1})}]$ relies inversely on bandwidth $h$. This suggests that we shall take limit of $M$ before limit of $L$, otherwise error bound in this prediction step will reach infinity before $M$ can constrain it.
\vspace{18pt}

\subsection{Bayesian update convergence analysis}\label{sec43}

In Bayesian update step, we have:
$$\overline{Y}_{k}^{i,M}=\frac{p(O_{t_{k}}|S_{k}=\widetilde{X}_{k}^{i})\overline{Y}_{k}^{O_{t_{k-1}},i,M}}{\sum_{j=1}^{N}p(O_{t_{k}}|S_{k}=\widetilde{X}_{k}^{j})\frac{1}{N}} \quad i=1,\dots,N$$

where $O_{t_{k}}|S_{k} \sim N(O_{t_{k-1}}+\int_{t_{k-1}}^{t_k}{S_t}dt|S_{k},t_{k}-t_{k-1})$\\

The corresponding empirical posterior density estimate is:
\begin{equation*}
\begin{aligned}
    \overline{Y}_{k}^{N,M}(x)=\overline{Y}_{k}^{i,M} \quad \mbox{if } \quad x=\widetilde{X}_{k}^{i}
\end{aligned}
\end{equation*}

\begin{lemma} \label{lemma431}

Assume $\lim_{M \rightarrow \infty}\overline{Y}_{k}^{O_{t_{k-1}},N,M}(x)=p(S_{k}=x|O_{t_{k-1}})$ pointwise with strong convergence rate $\frac{1}{\sqrt{M}}$ for RMSE, then for $\forall x \in R^{d_x}$:

$$\lim_{N \rightarrow \infty} \lim_{M \rightarrow \infty} \overline{Y}_{k}^{N,M}(x)=p(S_{k}=x|O_{t_{k}})$$

where $S_{k}$ is a discrete time update of $S_{t_{k}}$:
$$S_{k}=S_{t_{k-1}}+g(S_{t_{k-1}})(t_{k}-t_{k-1})+\sigma_{t_{k-1}}\Delta W_{t_{k-1}}$$

and $p(S_{k}=x|O_{t_{k}})$ is empirical posterior pdf for $S_{k}$:
$$p(S_{k}=x|O_{t_{k}})=\frac{p(O_{t_{k}}|S_{k}=x)p(S_{k}=x|O_{t_{k-1}})}{\int p(O_{t_{k}}|S_{k})p(S_{k}|O_{t_{k-1}})dS_{k}}$$

The uniform convergence rate of RMSE is $\frac{1}{\sqrt{M}}$ and $\frac{1}{\sqrt{N}}$.
\end{lemma}

\vspace{18pt}
Proof:

For $\forall$ x $\in$ $R^{d_{x}}$:
\begin{small}
\begin{equation*}
\begin{aligned}
&E[|\overline{Y}_{k}^{N,M}(x)-p(S_{k}=x|O_{t_{k}})|]\\
&=E[|\frac{p(O_{t_{k}}|S_{k}=x)\overline{Y}_{k}^{O_{t_{k-1}},N,M}(x)}{\sum_{j=1}^{N}p(O_{t_{k}}|S_{k}=\widetilde{X}_{k}^{j})\frac{1}{N}}-\frac{p(O_{t_{k}}|S_{k}=x)p(S_{k}=x|O_{t_{k-1}})}{\int p(O_{t_{k}}|S_{k})p(S_{k}|O_{t_{k-1}})dS_{k}}|]\\
&=E[\frac{p(O_{t_{k}}|S_{k}=x)}{\sum_{j=1}^{N}p(O_{t_{k}}|S_{k}=\widetilde{X}_{k}^{j})\frac{1}{N}\int p(O_{t_{k}}|S_{k})p(S_{k}|O_{t_{k-1}})dS_{k}}\\
&\quad (\overline{Y}_{k}^{O_{t_{k-1}},N,M}(x)\int p(O_{t_{k}}|S_{k})p(S_{k}|O_{t_{k-1}})dS_{k}-p(S_{k}=x|O_{t_{k-1}})\sum_{j=1}^{N}p(O_{t_{k}}|S_{k}=\widetilde{X}_{k}^{j})\frac{1}{N})]\\
% &\leq C*(E[|(\overline{Y}_{k}^{O_{t_{k-1}},M}(x)-p(S_{k}=x|O_{t_{k-1}}))\int p(O_{t_{k}}|S_{k})p(S_{k}|O_{t_{k-1}})dS_{k}|]\\
% &+E[|\overline{p}(S_{k}=x|O_{t_{k-1}})(\int p(O_{t_{k}}|S_{k})\overline{p}(S_{k}|O_{t_{k-1}})dS_{k}-\int p(O_{t_{k}}|S_{k})\overline{Y}_{k}^{O_{t_{k-1}},M}(S_{k})dS_{k}|]\\
% &+E[|\overline{p}(S_{k}=x|O_{t_{k-1}})(\int p(O_{t_{k}}|S_{k})\overline{Y}_{k}^{O_{t_{k-1}},M}(S_{k})dS_{k}-\sum_{j=1}^{N}p(O_{t_{k}}|S_{k}=\widetilde{X}_{k}^{j})\frac{1}{N})|])\\
% &+E[|p(S_{k}=x|O_{t_{k-1}})(\int p(O_{t_{k}}|S_{k})p(S_{k}|O_{t_{k-1}})dS_{k}-\sum_{j=1}^{N}p(O_{t_{k}}|S_{k}=\widetilde{X}_{k}^{j})\frac{1}{N})|])\\
&\leq \frac{p(O_{t_{k}}|S_{k}=x)}{\int p(O_{t_{k}}|S_{k})p(S_{k}|O_{t_{k-1}})dS_{k}}\sqrt{E[|\frac{1}{\sum_{j=1}^{N}p(O_{t_{k}}|S_{k}=\widetilde{X}_{k}^{j})\frac{1}{N}}|^{2}]}\\
&\quad *(\sqrt{E[|(\overline{Y}_{k}^{O_{t_{k-1}},N,M}(x)-p(S_{k}=x|O_{t_{k-1}}))\int p(O_{t_{k}}|S_{k})p(S_{k}|O_{t_{k-1}})dS_{k}|^2}]\\
&\quad +\sqrt{E[|p(S_{k}=x|O_{t_{k-1}})(\int p(O_{t_{k}}|S_{k})p(S_{k}|O_{t_{k-1}})dS_{k}-\sum_{j=1}^{N}p(O_{t_{k}}|S_{k}=\widetilde{X}_{k}^{j})\frac{1}{N})|^2])}\\
&\leq \frac{p(O_{t_{k}}|S_{k}=x)}{E[p(O_{t_{k}}|S_{k})]}O(\sqrt{\frac{Var_{t_{k}}^{x}[{Y_{k-1}^{N,M}(\widetilde{X}_{k-1})}]}{M}})+\frac{p(O_{t_{k}}|S_{k}=x)p(S_{k}=x|O_{t_{k-1}})}{E[p(O_{t_{k}}|S_{k})]^2}\sqrt{\frac{Var(p(O_{t_{k}}|S_{k}))}{N}}+O(\frac{1}{\sqrt{N}})\\
%%& \leq O(\sqrt{\frac{Var_{t_{k}}^{x}[{Y_{k-1}^{N,M}(\widetilde{X}_{k-1})}]}{M}})+\frac{C}{\sqrt{\Delta t^{d_x}}}\frac{1}{\sqrt{N}}+O(\frac{1}{\sqrt{N}})
%%& \leq \frac{p(O_{t_{k}}|S_{k}=x)}{E[p(O_{t_{k}}|S_{k})]}O(\frac{1}{\sqrt{M}})+\frac{p(O_{t_{k}}|S_{k}=x)\sqrt{E[p^{2}(O_{t_{k}}|S_{k}=x)]}}{E[p(O_{t_{k}}|S_{k})]^{2}}{\frac{p(S_{k}=x|O_{t_{k-1}})}{\sqrt{N}}}+O(\frac{1}{\sqrt{N}})
\end{aligned}
\end{equation*}
\end{small}

The third inequality comes from Holder's inequality. And to reach the next inequality, we define $\overline{S}=\sum_{j=1}^{N}p(O_{t_{k}}|S_{k}=\widetilde{X}_{k}^{j})\frac{1}{N}$, by Taylor expansion, we have:
$$\frac{1}{\overline{S}^2}=\frac{1}{E[\overline{S}]^2}-\frac{2}{E[\overline{S}]^3}(\overline{S}-E[\overline{S}])+\frac{3}{\epsilon^4}(\overline{S}-E[\overline{S}])^2$$

Taking expectation on both sides, we have $E[\frac{1}{\overline{S}^2}]=\frac{1}{E[\overline{S}]^2}+O(\frac{1}{N})$, and hence get the last inequality.

Moreover, both $p(O_{t_{k}}|S_{k})$ and $p(S_{k}=x|O_{t_{k-1}})$ are Gaussian densities and bounded, and since $O_{t_{k}}|S_{k} \sim N(O_{t_{k-1}}+\int_{t_{k-1}}^{t_k}{S_t}dt|S_{k},t_{k}-t_{k-1})$, expectation for $p(O_{t_{k}}|S_{k})$ is also a bounded positive constant, relying inversely on $\Delta t$:
$$E[p(O_{t_{k}}|S_{k})]=\int \frac{1}{\sqrt{{(2\pi \Delta t)}^{d_y}}}e^{-\frac{(O_{t_k}-\mu(O_{t_k}|S_k))^2}{2\Delta t}}p(S_k|O_{t_{k-1}})dS_k) \leq \frac{1}{\sqrt{{(2\pi \Delta t)}^{d_y}}} $$
% \begin{equation*}
% \begin{aligned}
% &E[p(O_{t_{k}}|S_{k})]\\
% &=E[\frac{1}{\sqrt{2\pi \Delta t}}e^{-\frac{(O_{t_{k}}-\mu(O_{t_{k}}|S_k))^2}{2\Delta t}}]\\
% &=E[\frac{1}{\sqrt{2\pi \Delta t}}e^{-\frac{(\Delta V_t)^2}{2\Delta t}}]\\
% &=\frac{1}{2\sqrt{\pi \Delta t}}
% \end{aligned}
% \end{equation*}

So $\overline{Y}_{k}^{N,M}(x)$ converges to $\overline{p}(S_{k}=x|O_{t_{k}})$ a.e. with strong uniform convergence rate $\frac{1}{M}$ and $\frac{1}{N}$ for mean squared error, and we need to take limit of $M$ before $N$, because error bound $Var_{t_{k}}^{x}[{Y_{k-1}^{N,M}(\widetilde{X}_{k-1})}]$ from prediction convergence in subsection \ref{sec42} relies inversely on kernel bandwidth $h$.

Timing factor $\Delta t$ from $p(O_{t_{k}}|S_{k})$ will cancel each other in numerator and denominator, but $\Delta t$ from $p(S_k|O_{t_{k-1}})$ will remain in the denominator, so we need to take limit of $\Delta t$  after $M$ and $N$, in case error bound from Bayesian update step explodes.
\vspace{18pt}

\subsection{FBSDE numerical scheme convergence}\label{sec44}

\begin{assumption}\label{assumption341}

Drift function $g(t,S_{t})$ and diffusion function $\sigma(t)$ satisfy Lipschitz and linear growth conditions: \\

(i) $E(|S_{0}|^{2})<\infty$\\

(ii) $|g(t,x)-g(t,y)|\leq C_{1}|x-y|$\\

(iii-1) $|g(t,x)|\leq C_{2}(1+|x|)$\\

(iii-2) $|g(t,x)|+|L^{0}g(t,x)|\leq C_{2}(1+|x|)$\\

\hspace{1cm} $|g(t,x)|+|L^{j}g(t,x)|\leq C_{2}(1+|x|)$\\

\hspace{1cm} where $j=1,\cdots,d_{w}$, $L^{0}g=\frac{\partial g}{\partial t}+J_{g}g$,  $L^{j}g=J_{g}\sigma^{j}(t)$\\

\hspace{1cm} $J_{g}$ is Jacobian matrix and $\sigma^{j}$ is $j_{th}$ column of diffusion matrix $\sigma$\\

(iv) $|g(s,x)-g(t,x)|+|\sigma(s)-\sigma(t)| \leq C_{3}(1+|x|)|s-t|^{1/2}$\\

(v) $\sigma(t) \in C_{b}^{1}([0,T])$ and $g(t,x) \in C_{b}^{1,3}([0,T] \times R^{d_{x}})$\\

where for vector v, $|v|=\sum_{i}|v_{i}|$, and for matrix m, $|m|=\sum_{i,j}|m_{i,j}|$\\

for $\forall s,t \in [0,T]$, $x,y \in R^{d_{x}}$ and  $C_{b}^{m,n}$ is space of functions with bounded, continuous derivatives up to order m in time and order n in space.\\

(iii-1) is assumption for EM scheme and (iii-2) is assumption for Milstein scheme.

\end{assumption}

\begin{lemma}

Under assumption \ref{assumption341}, $$\lim_{\Delta t \rightarrow 0}p(S_{k}=x|O_{t_{k}})=p(S_{t_{k}}=x|O_{t_{k}}) \quad a.e.$$

where $\displaystyle \Delta t=\max_{k=1,\cdots,K}t_{k}-t_{k-1}$

The uniform convergence rate of mean squared error is $\Delta t$ with assumption (iii-1) and $\Delta^{2} t$ with stronger assumption (iii-2).

\end{lemma}

\vspace{18pt}
Proof:

% Define $\int p(O_{t_{k}}|S_{k})p(S_{k}|O_{t_{k-1}})dS_{k}=C'_{1}=\frac{C_{1}}{\sqrt{{\Delta t}^{d_{y}}}}$ and $\int p(O_{t_{k}}|S_{t_{k}})p(S_{t_{k}}|O_{t_{k-1}})dS_{t_{k}}=C'_{2}=\frac{C_{2}}{\sqrt{{\Delta t}^{d_{y}}}}$.

% Since $p(O_{t_{k}}|S_{k})$ is Gaussian density, $C'_{1}$ and $C'_{2}$ are time dependent constants (we add superscript to indicate them as pseudo constants), after scaling them by $\sqrt{{\Delta t}^{d_{y}}}$, we can get time-independent constants $C_{1}$ and $C_{2}$.

% \begin{equation*}
% \begin{aligned}
% &|p(S_{k}|O_{t_{k}})-p(S_{t_{k}}|O_{t_{k}})|\\
% & =|\frac{p(O_{t_{k}}|S_{k})p(S_{k}|O_{t_{k-1}})}{C'_{1}}-\frac{p(O_{t_{k}}|S_{t_{k}})p(S_{t_{k}}|O_{t_{k-1}})}{C'_{2}}|\\
% & = \frac{1}{C'_{1}C'_{2}}|p(O_{t_{k}}|S_{k})p(S_{k}|O_{t_{k-1}})C'_{2}-p(O_{t_{k}}|S_{t_{k}})p(S_{t_{k}}|O_{t_{k-1}})C'_{1}|\\
% & \leq C_{3}\sqrt{(\Delta t)^{d_{y}-d_{x}}}|C'_{2}-C'_{1}|\\
% & + \frac{1}{C'_{2}}|p(O_{t_{k}}|S_{k})p(S_{k}|O_{t_{k-1}})-p(O_{t_{k}}|S_{t_{k}})p(S_{t_{k}}|O_{t_{k-1}})|
% \end{aligned}
% \end{equation*}

Define $\int p(O_{t_{k}}|S_{k})p(S_{k}|O_{0})dS_{k}=C_{1}$ and $\int p(O_{t_{k}}|S_{t_{k}})p(S_{t_{k}}|O_{0})dS_{t_{k}}=C_{2}$.

\begin{equation*}
\begin{aligned}
&|p(S_{k}|O_{t_{k}})-p(S_{t_{k}}|O_{t_{k}})|\\
& =|\frac{p(O_{t_{k}}|S_{k})p(S_{k}|O_{t_{k-1}})}{\int p(O_{t_{k}}|S_{k})p(S_{k}|O_{t_{k-1}})dS_{k}}-\frac{p(O_{t_{k}}|S_{t_{k}})p(S_{t_{k}}|O_{t_{k-1}})}{\int p(O_{t_{k}}|S_{t_{k}})p(S_{t_{k}}|O_{t_{k-1}})dS_{t_{k}}}|\\
& =|\frac{p(O_{t_{k}}|S_{k})p(S_{k}|O_{0})}{C_{1}}-\frac{p(O_{t_{k}}|S_{t_{k}})p(S_{t_{k}}|O_{0})}{C_{2}}|\\
& = \frac{1}{C_{1}C_{2}}|p(O_{t_{k}}|S_{k})p(S_{k}|O_{0})C_{2}-p(O_{t_{k}}|S_{t_{k}})p(S_{t_{k}}|O_{0})C_{1}|\\
& \leq C_{3}|C_{2}-C_{1}|\\
& + \frac{1}{C_{2}}|p(O_{t_{k}}|S_{k})p(S_{k}|O_{0})-p(O_{t_{k}}|S_{t_{k}})p(S_{t_{k}}|O_{0})|
\end{aligned}
\end{equation*}

Since $p(O_{t_{k}}|x)p(x|O_{0})$ is proportional to Gaussian density, it's Lipschitz continuous: 
$$|p(O_{t_{k}}|S_{k})p(S_{k}|O_{0})-p(O_{t_{k}}|S_{t_{k}})p(S_{t_{k}}|O_{0})|\leq L|S_{k}-S_{t_{k}}|$$

$S_{k}$ converges to $S_{t_{k}}$ at rate $O(\sqrt{\Delta t})$ with assumption (iii-1) in assumption \ref{assumption341} for EM scheme and $O(\Delta t)$ with assumption (iii-2) for Milstein scheme.

$C_{2}$ weakly converges to $C_{1}$, convergence rate depends on both regularity of drift and diffusion coefficients in SDE and test functions, test function $p(O_{t_{k}}|x)$ is Gaussian density, under assumption (v) in assumption \ref{assumption341}, convergence rate is $O({\Delta t})$. 

It's worth mentioning that if we use the second equality, conditioned time in conditional probability is $t_{k-1}$, both transition density and likelihood density will depend inversly on $\Delta t$, and this will make analysis unnecessarily tricky. So we extend conditioned time from $t_{k-1}$ to initial time $0$, and $t_{k}$ can be treated as a time independent with $\Delta t$, which conforms with the traditional notation $T$ in SDE convergence analysis.

Taking expectation on both sides:
\begin{equation*}
\begin{aligned}
E[|p(S_{k}|O_{t_{k}})-p(S_{t_{k}}|O_{t_{k}})|] &\leq C_{3}|C_{2}-C_{1}|+C_{4}E[|S_{k}-S_{t_{k}}|]\\
& =C_{3}O(\Delta t)+C_{4} O(\sqrt{\Delta t}) \quad EM\\
& or \quad C_{3}O(\Delta t)+C_{4} O(\Delta t) \quad Milstein
\end{aligned}
\end{equation*}

Therefore, $\forall x \in R_{d_{x}}$, $p(S_{k}=x|O_{t_{k}})$ converges to $p(S_{t_{k}}=x|O_{t_{k}})$, with uniform convergence rate $\sqrt{\Delta t}$ or $\Delta t$ under different drift and diffusion coefficients regularities. 

\vspace{18pt}

\subsection{Unified Convergence Analysis}\label{sec45}
Combining all convergence results from previous subsections, we can propose following lemma.

\begin{lemma}\label{lemma451}

Under assumption $Y_{k-1}^{N,M}(x)=p(S_{t_{k-1}}=x|O_{t_{k-1}}) \quad a.e.$ and corresponding assumptions in subsections \ref{sec41}-\ref{sec44}, we have:
$$\lim_{\Delta t \rightarrow 0}\lim_{L \rightarrow \infty}\lim_{M \rightarrow \infty}Y_{k}^{N,M}(x)=p(S_{t_{k}}=x|O_{t_{k}}) \quad a.e.$$
with sufficiently large number of gradient descent steps in KDE learning.

% where $\Delta t=\frac{T}{K}$, $M$ is size of samples used to estimate expectation $E_{t_{k}}^{\widetilde{X}_{k}}$.

% L is kernel number in kernel learning. When L goes to infinity, so does global sample size N for state variable.

Uniform convergence rate for RMSE is $L^{-\frac{2}{4+d_x}}$, $\frac{1}{\sqrt{M}}$ and $\sqrt{\Delta t}$ or $\Delta t$.
\end{lemma}

\vspace{18pt}
Proof:

From prediction convergence in subsection \ref{sec42} and Bayesian update convergence in subsection \ref{sec43}, we can reach:
$$ \lim_{N \rightarrow \infty} \lim_{M \rightarrow \infty} \overline{Y}_{k}^{N,M}(x)=p(S_{k}=x|O_{t_{k}}) \quad a.e.$$

Combined with KDE kernel learning convergence in subsection \ref{sec41} , we can conclude that with sufficiently large number of gradient descent steps:
$$\lim_{L \rightarrow \infty} \lim_{M \rightarrow \infty} Y_{k}^{N,M}(x)=p(S_{k}=x|O_{t_{k}}) \quad a.e.$$

Finally, take $\lim_{\Delta t \rightarrow 0}$ on left side and apply FBSDE numerical scheme convergence result in subsection \ref{sec44}, we can derive final result:
$$\lim_{\Delta t \rightarrow 0}\lim_{L \rightarrow \infty}\lim_{M \rightarrow \infty}Y_{k}^{N,M}(x)=p(S_{t_{k}}=x|O_{t_{k}}) \quad a.e.$$

Convergence rate for RMSE is $L^{-\frac{2}{4+d_x}}$(KDE), $\frac{1}{\sqrt{M}}$ and $\sqrt{\Delta t}$(Euler) or $\Delta t$(Milstein). And since all error bounds are independent of x, convergence is uniform.

For limit sequence, we need to take limit for $L$ after $M$ since prediction error bounds depend inversely on kernel bandwidth $h$, and then take limit for $\Delta t$ after $L$ as time fraction is in the denominator of Bayesian update error bounds.

\vspace{18pt}

We start our convergence analysis in previous subsections with assumption $Y_{k-1}^{N,M}(x)=p(S_{t_{k-1}}=x|O_{t_{k-1}}) \quad a.e.$ at time $t_{k-1}$, this simplifies notations and facilities convergence analysis in different steps.

Next, we relax this assumption to the general case $\lim_{\Delta t \rightarrow 0}\lim_{L \rightarrow \infty}\lim_{M \rightarrow \infty}Y_{k-1}^{N,M}(x)=p(S_{t_{k-1}}=x|O_{t_{k-1}}) \quad a.e.$ at time $t_{k-1}$ and show that we can still reach same convergence result at time $t_{k}$ and hence conclude our convergence analysis across time. 

% This entails cumbersome notation differentiation between time $t_{k-1}$ and $t_{k}$, and therefore we unify analysis of effects of this general assumption on steps in previous subsections in this unified subsection.

\begin{lemma}\label{lemma452}

Under assumption $\lim_{\Delta t \rightarrow 0}\lim_{L \rightarrow \infty}\lim_{M \rightarrow \infty}Y_{k-1}^{N,M}(x)=p(S_{t_{k-1}}=x|O_{t_{k-1}}) \quad a.e.$ and corresponding assumptions in subsections \ref{sec41}-\ref{sec44}, we have:
$$\lim_{\Delta t \rightarrow 0}\lim_{L \rightarrow \infty}\lim_{M \rightarrow \infty}Y_{k}^{N,M}(x)=p(S_{t_{k}}=x|O_{t_{k}}) \quad a.e.$$
with sufficiently large number of gradient descent steps in KDE learning.

Local uniform convergence rate for RMSE is $L^{-\frac{2}{4+d_x}}$, $\frac{1}{\sqrt{M}}$ and $\sqrt{\Delta t}$ or $\Delta t$. 

Global convergence requires more stringent conditions on likelihood density, and one sufficient condition is $\sup\limits_{t,x} \frac{E^{x}_{t^+}[p(O_{t}|S_{t})]}{E[p(O_{t}|S_{t})]} <\frac{1}{2\sqrt{(1+T^2G^2)}}$, under which global convergence is also uniform  with same convergence rate as local convergence.
\end{lemma}

\vspace{18pt}
Proof:

To facilitate analysis, we differentiate notations of $\Delta t,M,N,L$ between time $t_{k-1}$ and $t_k$. Now assumption can be rewritten as:
$$\lim_{\Delta t_{k-1} \rightarrow 0}\lim_{L_{t_{k-1}} \rightarrow \infty}\lim_{M_{t_{k-1}} \rightarrow \infty} Y_{k-1}^{N_{t_{k-1}},M_{t_{k-1}}}(x)=p(S_{t_{k-1}}=x|O_{t_{k-1}}) \quad a.e.$$.

And we need to analyze the effect of assumption changes on substeps in section \ref{sec41}-\ref{sec44}.

Firstly, assumption change has no impact on KDE kernel learning convergence analysis in section \ref{sec41}, we can claim that with sufficiently large number of gradient descent steps:
% $$\lim_{L_{t_{k}} \rightarrow \infty}Y_{k}^{N_{t_{k}}}(x)=\lim_{N_{t_{k}} \rightarrow \infty}\overline{Y}_{k}^{N_{t_{k}},M_{t_{k}}}(x) \quad \forall x \in R^{d_{x}}$$

$$E[|Y_{k}^{N_{t_{k}},M_{t_{k}}}(x)-\overline{Y}_{k}^{N_{t_{k}},M_{t_{k}}}(x)|] \leq O(L_{t_{k}}^{-\frac{2}{4+d_x}}) \quad \forall x \in R^{d_x}, M \in R$$

% Substituting Bayesian update convergence results with KDE convergence results, we can reach an iterated limit $\lim_{L_{t_{k-1}} \rightarrow \infty}\lim_{L_{t_{k}} \rightarrow \infty}$, since $L_{t_{k}},L_{t_{k-1}}$ depend on $N_{t_{k}},N_{t_{k-1}}$ respectively and $N_{t_{k}}$ is independent with $N_{t_{k-1}}$, $L_{t_{k}},L_{t_{k-1}}$ are independent and can be combined into simultaneous limit:
% $$\lim_{\Delta t_{k-1} \rightarrow 0}\lim_{M_{t_{k-1}},M_{t_{k}} \rightarrow \infty}\lim_{L_{t_{k-1},L_{t_{k}}} \rightarrow \infty} Y_{k}^{N_{t_{k}}}(x)=p(S_{k}=x|O_{t_{k-1}}) \quad a.e.$$

Next, following prediction convergence analysis in section \ref{sec42}, we need to differentiate $Y_{k-1}^{N_{t_{k-1}},M_{t_{k-1}}}(x)$ from $p(S_{t_{k-1}}=x|O_{t_{k-1}})$.

\begin{small}
\begin{equation*}
\begin{aligned}
E&[(\overline{Y}_{k}^{O_{t_{k-1}},N_{t_{k}},M_{t_{k}}}(x)-p(S_{k}=x|O_{t_{k-1}}))^2]\\
&=E[((E_{t_{k}}^{x,M_{t_{k}}}[Y_{k-1}^{N_{t_{k-1}},M_{t_{k-1}}}(\widetilde{X}_{k-1})]-E_{t_{k}}^{x}[Y_{k-1}^{N_{t_{k-1}},M_{t_{k-1}}}(\widetilde{X}_{k-1})])\\
&+ (E_{t_{k}}^{x}[Y_{k-1}^{N_{t_{k-1}},M_{t_{k-1}}}(\widetilde{X}_{k-1})]-E_{t_{k}}^{x}[p(S_{t_{k-1}}=\widetilde{X}_{k-1}|O_{t_{k-1}})])\\
&-(E_{t_{k}}^{x,M_{t_{k}}}[\sum_{j=1}^{d_{x}}\frac{\partial g_{j}}{\partial x_{j}}(\widetilde{X}_{k-1})Y_{k-1}^{N_{t_{k-1}},M_{t_{k-1}}}(\widetilde{X}_{k-1})]\Delta t_{k-1}-E_{t_{k}}^{x}[\sum_{j=1}^{d_{x}}\frac{\partial g_{j}}{\partial x_{j}}(\widetilde{X}_{k-1})Y_{k-1}^{N_{t_{k-1}},M_{t_{k-1}}}(\widetilde{X}_{k-1})]\Delta t_{k-1})])\\
&-(E_{t_{k}}^{x}[\sum_{j=1}^{d_{x}}\frac{\partial g_{j}}{\partial x_{j}}(\widetilde{X}_{k-1})Y_{k-1}^{N_{t_{k-1}},M_{t_{k-1}}}(\widetilde{X}_{k-1})]\Delta t_{k-1})]-E_{t_{k}}^{x}[\sum_{j=1}^{d_{x}}\frac{\partial g_{j}}{\partial x_{j}}(\widetilde{X}_{k-1})p(S_{t_{k-1}}=\widetilde{X}_{k-1}|O_{t_{k-1}})]\Delta t_{k-1}))^2]\\
&\leq 4E[(E_{t_{k}}^{x,M_{t_{k}}}[Y_{k-1}^{N_{t_{k-1}},M_{t_{k-1}}}(\widetilde{X}_{k-1})]-E_{t_{k}}^{x}[Y_{k-1}^{N_{t_{k-1}},M_{t_{k-1}}}(\widetilde{X}_{k-1})])^2]\\
& + 4(E_{t_{k}}^{x}[Y_{k-1}^{N_{t_{k-1}},M_{t_{k-1}}}(\widetilde{X}_{k-1})]-E_{t_{k}}^{x}[p(S_{t_{k-1}}=\widetilde{X}_{k-1}|O_{t_{k-1}})])^2\\
&+ 4E[(E_{t_{k}}^{x,M_{t_{k}}}[\sum_{j=1}^{d_{x}}\frac{\partial g_{j}}{\partial x_{j}}(\widetilde{X}_{k-1})Y_{k-1}^{N_{t_{k-1}},M_{t_{k-1}}}(\widetilde{X}_{k-1})]\Delta t_{k-1}-E_{t_{k}}^{x}[\sum_{j=1}^{d_{x}}\frac{\partial g_{j}}{\partial x_{j}}(\widetilde{X}_{k-1})Y_{k-1}^{N_{t_{k-1}},M_{t_{k-1}}}(\widetilde{X}_{k-1})]\Delta t_{k-1})^2]\\
&+ 4(E_{t_{k}}^{x}[\sum_{j=1}^{d_{x}}\frac{\partial g_{j}}{\partial x_{j}}(\widetilde{X}_{k-1})Y_{k-1}^{N_{t_{k-1}},M_{t_{k-1}}}(\widetilde{X}_{k-1})]\Delta t_{k-1}-E_{t_{k}}^{x}[\sum_{j=1}^{d_{x}}\frac{\partial g_{j}}{\partial x_{j}}(\widetilde{X}_{k-1})p(S_{t_{k-1}}=\widetilde{X}_{k-1}|O_{t_{k-1}})]\Delta t_{k-1})^2\\
& \leq 4\frac{Var_{t_{k}}^{x}[Y_{k-1}^{N_{t_{k-1}},M_{t_{k-1}}}(\widetilde{X}_{k-1})]}{M_{t_{k}}}+4T^2\frac{Var_{t_{k}}^{x}[\sum_{j=1}^{d_{x}}\frac{\partial g_{j}}{\partial x_{j}}(\widetilde{X}_{k-1})Y_{k-1}^{N_{t_{k-1}},M_{t_{k-1}}}(\widetilde{X}_{k-1})]}{M_{t_{k}}}\\
&+ 4(E_{t_{k}}^{x}[Y_{k-1}^{N_{t_{k-1}},M_{t_{k-1}}}(\widetilde{X}_{k-1})-p(S_{t_{k-1}}=\widetilde{X}_{k-1}|O_{t_{k-1}})])^2\\
&+ 4T^2(E_{t_{k}}^{x}[\sum_{j=1}^{d_{x}}\frac{\partial g_{j}}{\partial x_{j}}(\widetilde{X}_{k-1})(Y_{k-1}^{N_{t_{k-1}},M_{t_{k-1}}}(\widetilde{X}_{k-1})-p(S_{t_{k-1}}=\widetilde{X}_{k-1}|O_{t_{k-1}}))])^2\\
& \leq O(\frac{1}{M_{t_{k}}})+4(1+T^2G^2)E_{t_{k}}^{x}[Y_{k-1}^{N_{t_{k-1}},M_{t_{k-1}}}(\widetilde{X}_{k-1})-p(S_{t_{k-1}}=\widetilde{X}_{k-1}|O_{t_{k-1}})]^2
\end{aligned}
\end{equation*}
\end{small}

% Since $Y_{k-1}^{N_{t_{k-1}}}(x)$ and $p(S_{t_{k-1}}=x|O_{t_{k-1}})$ are Gaussian densities and bounded, and $|\sum_{j=1}^{d_{x}}\frac{\partial g_{j}}{\partial x_{j}}|$ is also assumed to be bounded, applying DCT, we can conclude that:
% $$\lim_{M_{t_{k}} \rightarrow \infty}\lim_{\Delta t_{k-1} \rightarrow 0}\lim_{M_{t_{k-1}} \rightarrow \infty}\lim_{L_{t_{k-1}} \rightarrow \infty} \overline{Y}_{k}^{O_{t_{k-1}},M_{t_{k}}}(x)=p(S_{k}=x|O_{t_{k-1}}) \quad a.e.$$

% Error bound items with regard to $M_{t_{k}}$ are independent with those related with $\Delta t_{k-1}, M_{t_{k-1}}, L_{t_{k-1}}$, therefore we can move $\lim_{M_{t_{k}}\rightarrow \infty}$ after $\lim_{M_{t_{k-1}}\rightarrow \infty}$, and combine iterated limits into simultaneous limits:
% $$\lim_{\Delta t_{k-1} \rightarrow 0}\lim_{M_{t_{k-1}},M_{t_{k}} \rightarrow \infty}\lim_{L_{t_{k-1}} \rightarrow \infty} \overline{Y}_{k}^{O_{t_{k-1}},M_{t_{k}}}(x)=p(S_{k}=x|O_{t_{k-1}}) \quad a.e.$$

Then, as in Bayesian update step in section \ref{sec43}, we apply same deduction and can reach result:
\begin{equation*}
\begin{aligned}
& E[|\overline{Y}_{k}^{N_{t_{k}},M_{t_{k}}}(x)-p(S_{k}=x|O_{t_{k}})|] \\
% &\leq \frac{2p(O_{t_{k}}|S_{k}=x)}{\int p(O_{t_{k}}|S_{k})p(S_{k}|O_{t_{k-1}})dS_{k}}\sqrt{E[|\frac{1}{\sum_{j=1}^{N}p(O_{t_{k}}|S_{k}=\widetilde{X}_{k}^{j})\frac{1}{N}}|^{2}]}\\
% &\quad *(\sqrt{E[|(\overline{Y}_{k}^{O_{t_{k-1}},M}(x)-p(S_{k}=x|O_{t_{k-1}}))\int p(O_{t_{k}}|S_{k})p(S_{k}|O_{t_{k-1}})dS_{k}|^2}]\\
% &\quad +\sqrt{E[|p(S_{k}=x|O_{t_{k-1}})(\int p(O_{t_{k}}|S_{k})p(S_{k}|O_{t_{k-1}})dS_{k}-\sum_{j=1}^{N}p(O_{t_{k}}|S_{k}=\widetilde{X}_{k}^{j})\frac{1}{N})|^2])}\\
&\leq \frac{p(O_{t_{k}}|S_{k}=x)}{E[p(O_{t_{k}}|S_{k})]}\sqrt{E[|\overline{Y}_{k}^{O_{t_{k-1}},N_{t_{k}},M_{t_{k}}}(x)-p(S_{k}=x|O_{t_{k-1}})|^2]}+O(\frac{1}{\sqrt{N_{t_{k}}}})\\
& \leq \frac{2\sqrt{(1+T^2G^2)}}{\sqrt{(2\pi\Delta t_k)^{d_y}}E[p(O_{t_{k}}|S_{k})]}E_{t_{k}}^{x}[|Y_{k-1}^{N_{t_{k-1}},M_{t_{k-1}}}(\widetilde{X}_{k-1})-p(S_{t_{k-1}}=\widetilde{X}_{k-1}|O_{t_{k-1}})|]\\
&+O(\frac{1}{\sqrt{M_{t_{k}}}})+O(\frac{1}{\sqrt{N_{t_{k}}}})
\end{aligned}
\end{equation*}

% Error bound items with $N_{t_{k}}$ are independent with all other errors, $\lim_{N_{t_{k}}\rightarrow \infty}$ can be put in any location in sequence of iterated limits, and we place it in rightmost position to facilitate applying KDE convergence results: 
% $$\lim_{\Delta t_{k-1} \rightarrow 0}\lim_{M_{t_{k-1}},M_{t_{k}} \rightarrow \infty}\lim_{L_{t_{k-1}} \rightarrow \infty}\lim_{N_{t_{k}} \rightarrow \infty} \overline{Y}_{k}^{N_{t_{k}},M_{t_{k}}}(x)=p(S_{k}=x|O_{t_{k-1}}) \quad a.e.$$

Finally, FBSDE numerical scheme convergence analysis in section \ref{sec44} is also unaffected:
\begin{equation*}
\begin{aligned}
E[|p(S_{k}|O_{t_{k}})-p(S_{t_{k}}|O_{t_{k}})|] < &O(\sqrt{\Delta t_{k}}) \quad EM\\
& or\quad O(\Delta t_{k}) \quad Milstein
\end{aligned}
\end{equation*}

% Error bound items with $\Delta t_{k}$ are independent with all other error items, so we can take $\lim_{\Delta t_{k} \rightarrow 0}$ on left side and combine $\Delta t_{k}$ with $\Delta t_{k-1}$ into simultaneous limit:
% $$\lim_{\Delta t_{k-1},\Delta t_{k} \rightarrow 0}\lim_{M_{t_{k-1}},M_{t_{k}} \rightarrow \infty}\lim_{L_{t_{k-1},L_{t_{k}}} \rightarrow \infty} Y^{N_{t_{k}}}(x)=p(S_{t_{k}}=x|O_{t_{k-1}}) \quad a.e.$$

Combining error bounds in all previous substeps under the new assumption, we can conclude that:
\begin{equation*}
\begin{aligned}
&E[|Y_k^{N_{t_{k}},M_{t_{k}}}(x)-p(S_{t_{k}}=x|O_{t_{k}})|] \\
& \leq E[|Y_k^{N_{t_{k}},M_{t_{k}}}(x)-\overline{Y}_{k}^{N_{t_{k}},M_{t_{k}}}(x)|]+E[|\overline{Y}_{k}^{N_{t_{k}},M_{t_{k}}}(x)-p(S_{k}=x|O_{t_{k}})|]\\
& \quad +E[|p(S_{k}=x|O_{t_{k}})-p(S_{t_{k}}=x|O_{t_{k}})|]\\
& \leq \frac{2\sqrt{(1+T^2G^2)}p(O_{t_{k}}|S_{k}=x)}{E[p(O_{t_{k}}|S_{k})]}E_{t_{k}}^{x}[E[|Y_{k-1}^{N_{t_{k-1}},M_{t_{k-1}}}(\widetilde{X}_{k-1})-p(S_{t_{k-1}}=\widetilde{X}_{k-1}|O_{t_{k-1}})|]]\\
& +O(L_{t_{k}}^{-\frac{2}{4+d_x}})+O(\frac{1}{\sqrt{M_{t_{k}}}})+O(\sqrt{\Delta t_{k}})(or\,\,O(\Delta t_{k})) \quad{(a)}\\
& \leq \frac{2\sqrt{(1+T^2G^2)}}{\sqrt{(2\pi\Delta t_k)^{d_y}}E[p(O_{t_{k}}|S_{k})]}E[|Y_{k-1}^{N_{t_{k-1}},M_{t_{k-1}}}(\widetilde{X}_{k-1})-p(S_{t_{k-1}}=\widetilde{X}_{k-1}|O_{t_{k-1}})|]\\
& +O(L_{t_{k}}^{-\frac{2}{4+d_x}})+O(\frac{1}{\sqrt{M_{t_{k}}}})+O(\sqrt{\Delta t_{k}})(or\,\,O(\Delta t_{k})) \quad{(b)}
\end{aligned}
\end{equation*}

In the above equation, we further extend expectation to incorporate variables at time $t_{k-1}$, and hence we incur an additional expectation in the third equality (a) and form an recurrence relationship between errors across time. The last inequality (b) comes from the fact that when we bound error items independent of x, conditional expectation $E_{t_{k}}^{x}$ is equivalent to unconditional expectation and can be omitted.

Taking sequential limits on both sides, we have:
$$\lim_{\Delta t_{k} \rightarrow 0} \lim_{L_{k} \rightarrow \infty}\lim_{M_{k} \rightarrow \infty}\lim_{\Delta t_{k-1} \rightarrow 0}\lim_{L_{k-1} \rightarrow \infty}\lim_{M_{k-1} \rightarrow \infty}Y_{k}^{N_{t_{k}},M_{t_{k}}}(x)=p(S_{t_{k}}=x|O_{t_{k}}) \quad a.e.$$

Since error bounds related with $\Delta t,M,N,L$ are independent between time $t_{k-1}$ and $t_{k}$, we can first swap original sequential limits into:
$$\lim_{\Delta t_{k} \rightarrow 0}\lim_{\Delta t_{k-1} \rightarrow 0}\lim_{L_{k} \rightarrow \infty}\lim_{L_{k-1} \rightarrow \infty}\lim_{M_{k} \rightarrow \infty}\lim_{M_{k-1} \rightarrow \infty}Y_{k}^{N_{t_{k}},M_{t_{k}}}(x)=p(S_{t_{k}}=x|O_{t_{k}}) \quad a.e.$$

and then combine iterated limits into simultaneous limits:
$$\lim_{\Delta t_{k-1},\Delta t_{k} \rightarrow 0}\lim_{L_{t_{k-1},L_{t_{k}}} \rightarrow \infty}\lim_{M_{t_{k-1}},M_{t_{k}} \rightarrow \infty} Y_{k}^{N_{t_{k}},M_{t_{k}}}(x)=p(S_{t_{k}}=x|O_{t_{k-1}}) \quad a.e.$$

Taking $\Delta t, M,L,N$ to be equal across all time steps, we reach the final result:
$$\lim_{\Delta t \rightarrow 0}\lim_{L \rightarrow \infty}\lim_{M \rightarrow \infty}Y_{k}^{N,M}(x)=p(S_{t_{k}}=x|O_{t_{k}}) \quad a.e.$$

Same as lemma \ref{lemma451}, we need to maintain limit sequence order for $M$, $L$ and $\Delta t$.
% But we can exchange limit sequence among time, as long as we keep relative order between $\Delta t$, $M$ and $N$ during single time interval.

To analyze global convergence, we set recurrence coefficient R to be sample and time independent from inequality $(b)$: $R=\sup\limits_{k} \frac{2\sqrt{(1+T^2G^2)}}{\sqrt{(2\pi \Delta t_k)^{d_y}}E[p(O_{t_{k}}|S_{k})]}$, and R is time independent since ${\Delta t_k}^{d_y}$ in $\sqrt{(2\pi \Delta t_k)^{d_y}}$ and $E[p(O_{t_{k}}|S_{k})]$ will cancel each other.

By recurrence, we can derive:
$$E[|Y_k^{N,M}(x)-p(S_{t_{k}}|O_{t_{k}})|] \leq \frac{1-R^{k}}{1-R}[O(L^{-\frac{2}{4+d_x}})+O(\frac{1}{\sqrt{M}})+O(\sqrt{\Delta t})(or\,\,O(\Delta t))]$$

Therefore, as long as recurrence coefficient $R$ is smaller than 1, global error will converge. Unfortunately, $R$ is larger than $2\sqrt{(1+T^2G^2)}$. And the main reason that this recurrence coefficient fails to guarantee global convergence is that our upper error bound is too loose.

Following same argument, we now re-select an recurrence coefficient R from inequality $(a)$: $R=2\sqrt{(1+T^2G^2)}\sup\limits_{t,x} \frac{E^{x}_{t^+}[p(O_{t}|S_{t})]}{E[p(O_{t}|S_{t})]}$. The numerator is expectation of likelihood density propagated backward and the denominator is expectation of likelihood density propagated forward. When supreme of their ratio is smaller than $\frac{1}{2\sqrt{(1+T^2G^2)}}$, recurrence coefficient is kept below 1 and global convergence can be obtained with same convergence rate as local convergence.

The proposed recurrence coefficient R is only one sufficient condition for global convergence, when recurrence coefficient is sample or time dependent, global convergence may also be reached, but it's difficult to generalize those scenarios. What we can argue from inequality $(a)$ is global convergence relies heavily on properties of likelihood density, intuitively speaking, likelihood density propagated backward shall be expected to be smaller than likelihood density propagated forward for sufficient periods of time.

The property that recurrence coefficient is smaller than 1, at least for sufficient periods of time, is crucial for global convergence. In sequential Monte Carlo convergence analysis ([20]), Crisan and Doucet argue that error constant from time $c_t$ is independent of $N$ and hence convergence rate is $\frac{1}{N}$, and our previous analysis has suggested that $c_t$ can be unbounded when $\Delta t$ goes to 0 and global convergence will fail in that case. 
\vspace{18pt}

Putting together lemma \ref{lemma451} and \ref{lemma452}, we can extend convergence result for any maturity T.
\begin{theorem}
Under corresponding assumptions in subsections \ref{sec41}-\ref{sec44} and $\sup\limits_{t,x} \frac{E^{x}_{t^+}[p(O_{t}|S_{t})]}{E[p(O_{t}|S_{t})]}<\frac{1}{2\sqrt{(1+T^2G^2)}}$, for $\forall$ $T>0$, we have:

$$\lim_{\Delta t \rightarrow 0}\lim_{L \rightarrow \infty}\lim_{M \rightarrow \infty}Y_{T}^{N,M}(x)=p(S_{T}=x|O_{T}) \quad a.e.$$

with sufficiently large number of gradient descent steps in KDE learning.

Global uniform convergence rate for RMSE is $L^{-\frac{2}{4+d_x}}$, $\frac{1}{\sqrt{M}}$ and $\sqrt{\Delta t}$ or $\Delta t$. 
\end{theorem}

\vspace{18pt}
Proof:

At time 0, $Y_{0}^{N,M}(x)=p(S_{0}=x|O_{0})$, applying lemma \ref{lemma451}, we have:
$$\lim_{\Delta t \rightarrow 0}\lim_{L \rightarrow \infty}\lim_{M \rightarrow \infty}Y_{1}^{N,M}(x)=p(S_{t_{1}}=x|O_{t_{1}}) \quad a.e.$$

Now assumptions for lemma \ref{lemma452} are satisfied, we can apply this lemma recursively until maturity, under sufficient condition $\sup\limits_{t,x} \frac{E^{x}_{t^+}[p(O_{t}|S_{t})]}{E[p(O_{t}|S_{t})]}<\frac{1}{2\sqrt{(1+T^2G^2)}}$, global error is also shown to be uniform and finite, therefore we can claim:
$$\lim_{\Delta t \rightarrow 0}\lim_{L \rightarrow \infty}\lim_{M \rightarrow \infty}Y_{T}^{N,M}(x)=p(S_{T}=x|O_{T}) \quad a.e.$$

\section{Conclusions}
In this paper, we analyze the convergence of FBSDE filter. In each local time interval, we separate the implementation into prediction, Bayesian update and kernel learning steps, and analyze errors in each step along with discretization error. Local mean squared errors are proved to uniformly converge at rate of $L^{-\frac{4}{4+d_{x}}}$,$\frac{1}{M}$ and ${\Delta t}$. After accumulating local errors over time, global error will only converge when likelihood density satisfies certain conditions.

% %%%% Acknowledgments %%%%%%%%
% \section*{Acknowledgments}
% The author expresses thanks to the people helping with this work.
% Financial support also appears in this part, with grant number(s) following.
% The full title of each fund is required.

%%%% Bibliography  %%%%%%%%%%
%Please do not use BibTex.

\end{document}